\PassOptionsToPackage{hyperfootnotes=false}{hyperref}
\documentclass[10pt,a4paper]{article}
\usepackage{fontawesome5}
\usepackage{yalenlp}
\usepackage{adjustbox}
\usepackage{marvosym}

\usepackage{amsmath}
\usepackage{booktabs}
\usepackage{caption}
\usepackage{float}
\usepackage{graphicx}
\usepackage{xurl}

\usepackage{wrapfig}
\usepackage{listings}
\usepackage{microtype}
\usepackage{tabularx}
\usepackage{xcolor}
\usepackage{xspace}
\usepackage{longtable}

\definecolor{codeblue}{RGB}{35,90,155}
\definecolor{codegreen}{RGB}{30,120,85}
\definecolor{codegray}{RGB}{90,90,90}

\hypersetup{
  colorlinks=true,
  linkcolor=codeblue,
  citecolor=codegreen,
  urlcolor=codeblue
}

\renewcommand{\arraystretch}{1.08}

\newcommand{\cmark}{\textcolor{codegreen}{$\surd$}}
\newcommand{\xmark}{\textcolor{codegray}{$\times$}}
\newcommand{\systemname}{CodeActionBench\xspace}
\newcommand{\runcode}{\texttt{run\_code}\xspace}

\hypersetup{
  pdftitle={CodeActionBench: Evaluating Agentic Code-as-Policy for Embodied Manipulation},
  pdfauthor={Yiheng Lyu, Xueying Jiang, Wenhao Li, Shijian Lu, Gongjie Zhang},
  pdfsubject={Embodied manipulation benchmark},
  pdfkeywords={CodeActionBench, benchmark, robotics, Code-as-Policy}
}
\newcommand{\authornotes}{%
  \begingroup
  \renewcommand{\thefootnote}{\Letter}%
  \footnotetext{Corresponding authors: Shijian Lu and Gongjie Zhang.}%
  \renewcommand{\thefootnote}{\ensuremath{\dagger}}%
  \footnotetext{Project Leader.}%
  \endgroup
  \setcounter{footnote}{0}%
}
\AtBeginDocument{%
  \ifdefined\pdfminorversion
    \pdfminorversion=7
  \else
  \fi
}

\usepackage{latexml}
\renewcommand{\lablogo}{}
\definecolor{NTUNavy}{RGB}{24,30,61}
\definecolor{NTURed}{RGB}{190,20,48}
\definecolor{ResourcePurple}{HTML}{584B86}
\fancypagestyle{yalefirst}{%
  \fancyhf{}
  
  \fancyhead[L]{\includegraphics[height=0.9cm]{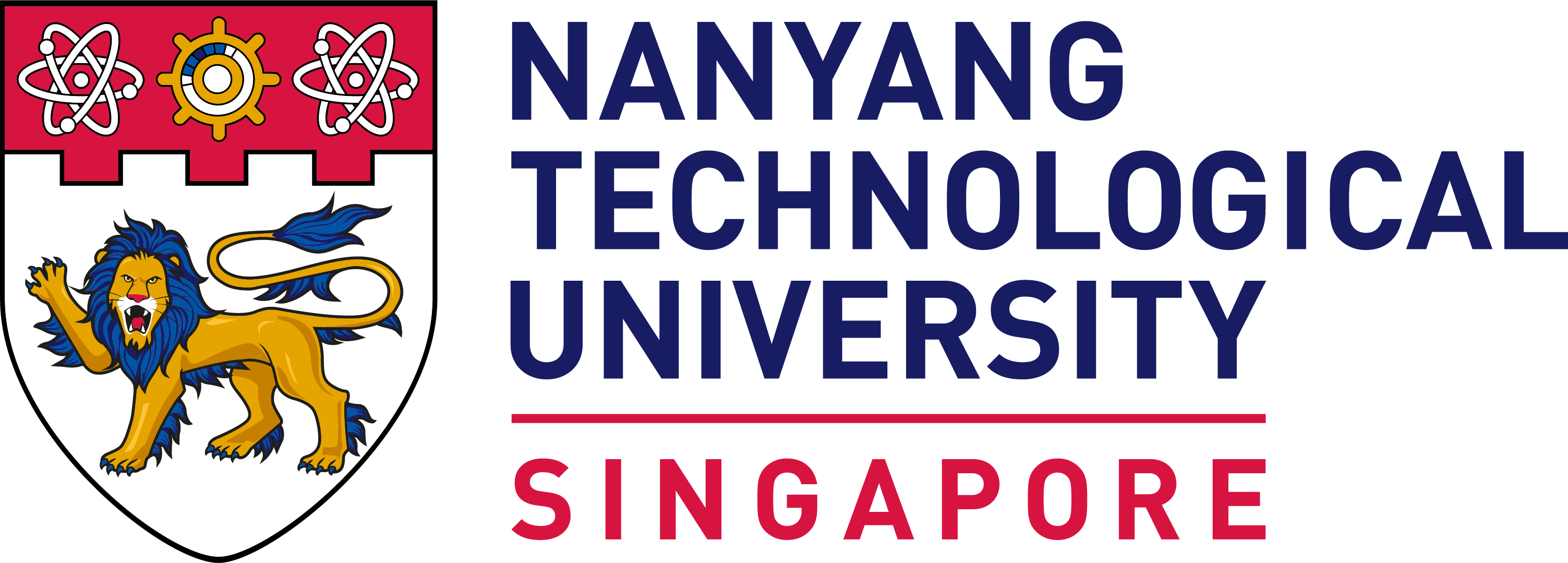}}
  \fancyhead[R]{\sffamily\scriptsize\color{NTUNavy}CODEACTIONBENCH}
  \fancyfoot[C]{\thepage}
}
\iflatexml
\usepackage{authblk}
\title{CodeActionBench: Evaluating Agentic Code-as-Policy for Embodied Manipulation}
\author[1]{Yiheng Lyu}
\author[1]{Xueying Jiang}
\author[1]{Wenhao Li}
\author[1]{Shijian Lu}
\author[2]{Gongjie Zhang\textsuperscript{\textdagger}}
\affil[1]{Nanyang Technological University, Singapore}
\affil[2]{Independent Researcher}
\else
\title{CodeActionBench: Evaluating Agentic\\Code-as-Policy for Embodied Manipulation}
\author{%
Yiheng Lyu\textsuperscript{1}\quad
Xueying Jiang\textsuperscript{1}\quad
Wenhao Li\textsuperscript{1}\quad
Shijian Lu\textsuperscript{1,\Letter}\quad
Gongjie Zhang\textsuperscript{2,\Letter,$\dagger$}\\[7pt]
\normalfont\small\textsuperscript{1}Nanyang Technological University, Singapore\\[2pt]
\normalfont\small\textsuperscript{2}Independent Researcher\\[6pt]
{\normalfont\small\sffamily
\href{https://github.com/lyhkk/CodeActionBench}{\textcolor{black}{\faGithub}\hspace{0.4em}\textcolor{ResourcePurple}{GitHub}}%
\hspace{1.7em}%
\href{https://codeactionbench.org/benchmark}{\raisebox{-0.14em}{\includegraphics[height=1.1em]{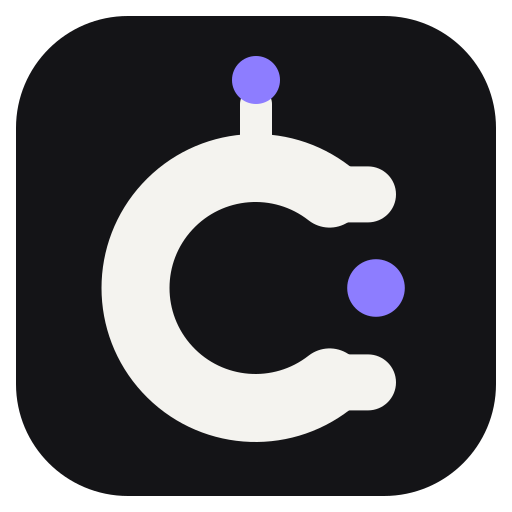}}\hspace{0.4em}\textcolor{ResourcePurple}{Project Website}}}
}
\fi
\date{}

\begin{document}
\maketitle
\iflatexml
\begin{center}
Corresponding authors: Shijian Lu and Gongjie Zhang.\\
\textdagger{} Project Leader: Gongjie Zhang.\\[4pt]
\href{https://github.com/lyhkk/CodeActionBench}{GitHub}\quad
\href{https://codeactionbench.org/benchmark}{Project Website}
\end{center}
\else
\thispagestyle{yalefirst}
\authornotes
\fi
\begin{abstract}
How well can general-purpose multimodal models turn visual understanding and reasoning into embodied manipulation via executable code? We introduce CodeActionBench, a benchmark of 25 manipulation tasks that evaluates this capability through agentic Code-as-Policy. Without task-specific fine-tuning, demonstrations, external specialist perception or grasp modules, privileged scene state, or predefined task policies, agents should select visual evidence, form task-relevant 3D estimates, construct manipulation targets, and iteratively execute and revise their policies. A shared robot API provides RGB observations, calibrated geometric operations, robot feedback, and bounded motion, leaving task-dependent decisions to the evaluated agent. Fixed task instances, resource budgets, and a hidden physical-outcome verifier support controlled comparisons across models and harness configurations. Extensive evaluations across nine configurations and 675 attempts achieve success rates ranging from 2.7\% to 73.3\%. The strongest configuration, GPT-6 Astra with Codex CLI, solves 22 of 25 tasks at least once in three attempts, demonstrating the best performance while still leaving substantial room for improvement. Trajectory analyses reveal difficulties in spatial alignment, object retention, and completion judgment, including task failures despite successfully completed motions. CodeActionBench provides a controlled testbed for measuring how general-purpose models translate their capabilities into manipulation behavior and for examining typical failure scenarios in that process.

\end{abstract}
\section{Introduction}
\label{sec:intro}

General-purpose multimodal models~\citep{openai2026astra,openai2026gpt56,anthropic2026opus5,anthropic2026sonnet5,deepmind2026gemini36flash,kimiteam2026kimik3,alibaba2026qwen38max,spacexai2026grok46} combine visual understanding, reasoning, and code generation within a single model. These capabilities suggest a compelling possibility for embodied manipulation: the same model could interpret a scene, determine how to interact with it, and express its decisions as executable robot behavior. Manipulation, however, requires these capabilities to work together under physical constraints. Recognizing an object is insufficient without locating a suitable contact point, constructing an actionable target, and responding when execution produces an unexpected outcome. How effectively general-purpose models can carry this process from visual evidence to successful manipulation remains an open evaluation question.

\begin{figure}[!t]
  \centering
  \includegraphics[width=\linewidth]{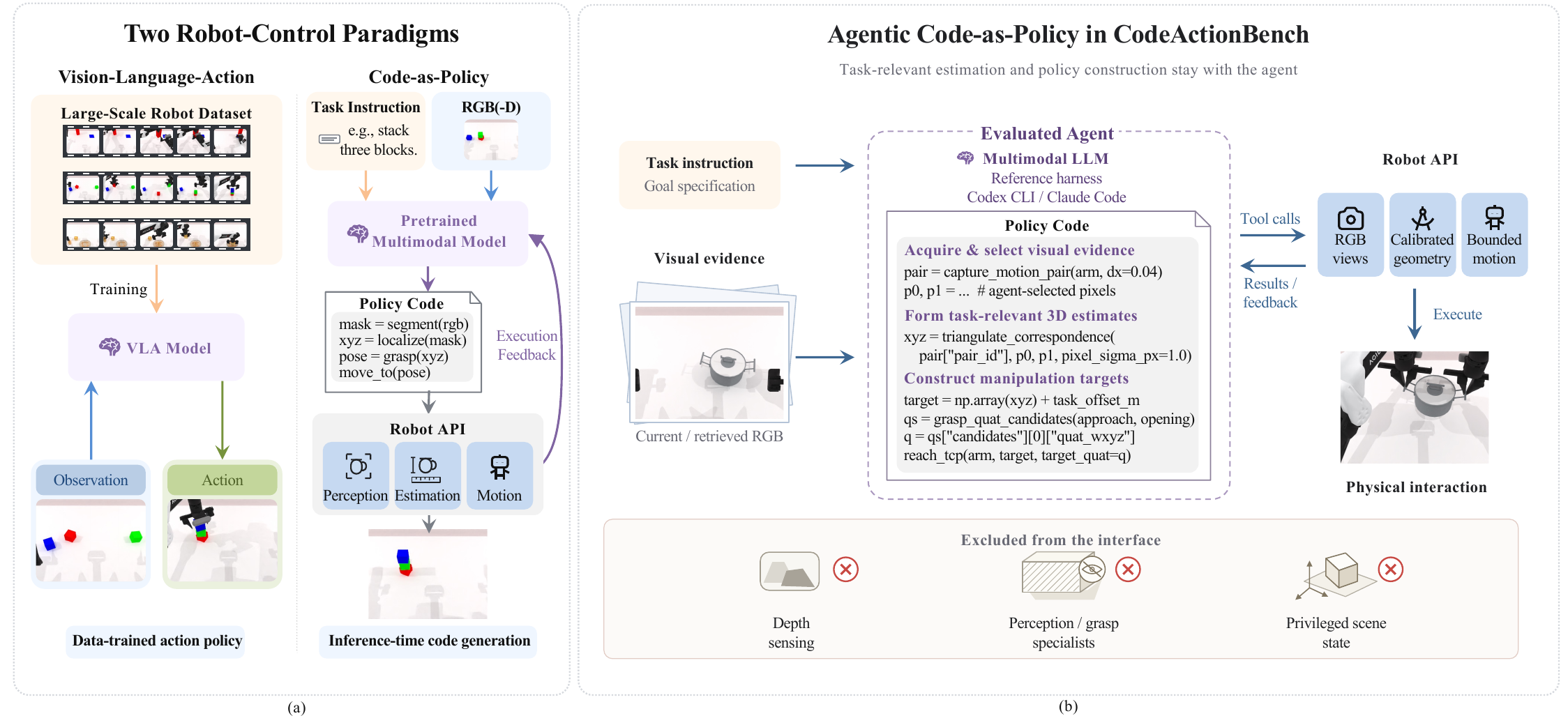}
  \caption{\textbf{Robot-control paradigms and agentic Code-as-Policy.} (a) VLA learns action policies from large-scale robot datasets; Code-as-Policy generates programs that compose robot APIs to perform tasks. (b) ~\systemname evaluates agents that autonomously construct task-relevant 3D estimates, manipulation targets, and policies, without depth sensing, external perception/grasp specialists, or privileged scene state.}
  \label{fig:overview}
\end{figure}

There are complementary ways to connect foundation-model capabilities to robot actions. Vision-language-action (VLA) approaches learn observation-to-action mappings through training on large-scale robot demonstration datasets \citep{zitkovich2023rt2,kim2025openvla,black2025pi0,black2025pi05,physicalintelligence2025pi06,li2026camvla,zhao2026geoprop}. OpenVLA~\citep{kim2025openvla}, for example, is trained on 970,000 real-world robot manipulation trajectories. Code-as-Policy offers another route: a general-purpose model constructs executable programs over robot APIs at inference time \citep{liang2023codeaspolicies}, and an agentic interaction loop allows it to inspect execution results and revise subsequent programs \citep{wang2024executable}. As illustrated in Figure~\ref{fig:overview}(a), this approach allows a pretrained model to apply its existing capabilities to manipulation through policy construction, without requiring additional task-specific training or demonstrations.

However, existing Code-as-Policy systems often supply object locations, grasp proposals, or task skills through external modules or privileged interfaces \citep{naouali2026vlcp,chen2026openeta,fu2026capx}. Their success therefore leaves open how well the general-purpose model can perform the perceptual reasoning and manipulation decisions supplied by those components. A multimodal model can both inspect images and generate code, making it possible to assign these responsibilities to the same model. Evaluating this possibility requires a setting in which task-dependent visual interpretation and policy construction remain with the evaluated model.

We introduce \systemname, a benchmark of 25 manipulation tasks designed around this requirement. The model should autonomously select visual evidence, construct task-relevant spatial estimates and manipulation targets, and decide how to act, recover, and stop. Auxiliary specialist perception and grasp outputs, privileged scene state, and predefined task policies are withheld. A shared robot API provides RGB observations, camera calibration, robot and contact feedback, geometric computations, and bounded motion. The model expresses its decisions through direct API calls or Python programs composing those calls, then uses execution feedback to revise its policy. Figure~\ref{fig:overview}(b) illustrates this division of responsibility: the model selects image correspondences and proposes grasp targets, while the API computes geometric quantities and checks candidate poses. Task-level planning and target selection remain with the model; inverse kinematics, motion planning, and low-level control remain shared infrastructure.

\begin{figure}[!t]
  \centering
  \includegraphics[width=\linewidth]{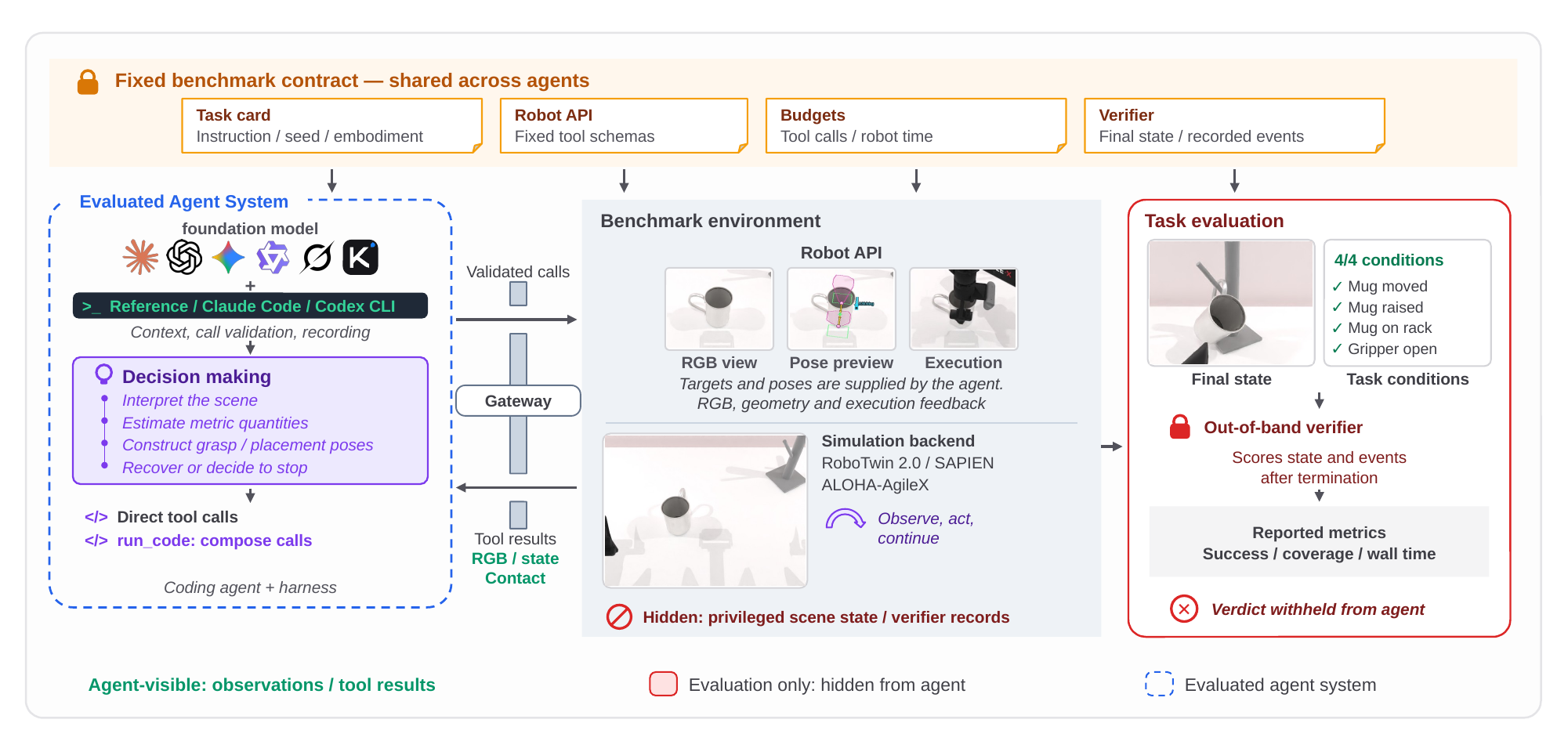}
  \caption{\textbf{Benchmark system and information boundaries.} Each evaluated agent comprises a model and its harness. Task instances, the robot API, resource budgets, and verification rules are fixed across configurations. A hidden verifier scores physical outcomes after termination without returning its verdict to the agent, separating task outcome from the agent's completion judgment.}
  \label{fig:system}
\end{figure}

\systemname turns this division of responsibility into a common evaluation protocol (see Figure~\ref{fig:system}). All configurations share the task instances, robot interface, resource budgets, and verification rules. A hidden verifier evaluates the final physical state and required task events without returning its verdict to the agent, allowing completion judgment to be assessed separately from task success. Recorded observations, programs, tool results, and physical consequences support trajectory-level analysis of how agents construct and execute their policies. This makes it possible to examine intermediate progress and unmet manipulation requirements that a binary success score alone cannot distinguish. Scores characterize model-and-harness configurations, with a shared reference harness supporting controlled comparisons among seven of the evaluated models.

We evaluate nine configurations across 675 attempts, without additional task-specific fine-tuning or demonstrations. Success rates range from 2.7\% to 73.3\%. The best-performing configuration, GPT-6 Astra with Codex CLI, solves 22 of 25 fixed task instances at least once in three attempts, while no configuration solves the entire suite. Trajectory analyses reveal difficulties in spatial alignment, object retention, and completion judgment. Failed tasks can contain successfully completed robot motions, highlighting the distinction between executing a commanded pose and establishing the physical relations required by the task. These findings demonstrate both the promise and the remaining limitations of general-purpose models undertaking visual interpretation and policy construction within the same embodied interaction loop.

Our contributions are threefold:
\begin{itemize}
\item We define a controlled setting for evaluating joint visual interpretation and policy construction by general-purpose multimodal models, without auxiliary specialist outputs, privileged scene state, or predefined task policies.
\item We introduce \systemname, a 25-task manipulation benchmark with a shared robot API, fixed resource budgets, hidden physical-outcome verification, and recorded trajectories for examining agent behavior.
\item We establish baselines across nine agent configurations and analyze their manipulation outcomes, policy execution, and completion judgments, revealing limitations beyond those captured by aggregate success rates.
\end{itemize}

\section{Related Work}
\label{sec:related}

\noindent\textbf{Vision-language-action policies.}
VLA models map visual observations and language instructions to robot actions with policies learned from robot data. RT-2~\citep{zitkovich2023rt2} represents actions as tokens, while RT-X~\citep{oneill2024openxembodiment}, Octo~\citep{ghosh2024octo}, and OpenVLA~\citep{kim2025openvla} train generalist policies on heterogeneous corpora. $\pi_{0.5}$~\citep{black2025pi05} further supports long-horizon manipulation. Recent work explores cross-embodiment transfer~\citep{luo2026beingh05,li2026dypesvla}, geometric inductive biases~\citep{peng2026g3vla}, in-context tool use~\citep{yang2026incontextvla}, and viewpoint robustness~\citep{li2026camvla}. However, transfer to new embodiments can require additional adaptation. \systemname instead evaluates agents that generate and revise executable policy code at inference time through a shared robot API.

\noindent\textbf{Code-as-Policy methods.}
Language-model robot planning incorporates affordances, execution feedback, and scene structure~\citep{ichter2023saycan,huang2023innermonologue,huang2023groundeddecoding,lin2023text2motion,rana2023sayplan}. Code as Policies~\citep{liang2023codeaspolicies} represents robot policies as programs that compose perception and control APIs. Related approaches explore programmatic task planning~\citep{singh2023progprompt,huang2023instruct2act}, code generation from demonstrations~\citep{wang2023demo2code}, integration with motion planning~\citep{chen2024roboscript,wang2024llm3}, and spatial constraints expressed through value maps or keypoint relations~\citep{huang2023voxposer,huang2025rekep}. CodeAct~\citep{wang2024executable} uses executable Python as an agent action space with execution feedback. \systemname supports both direct tool calls and Python programs that compose the same tools. Across these approaches, differences in the information and control functions provided by the interface make direct comparisons difficult.

\noindent\textbf{Code-as-Policy interfaces.}
Code-as-Policy evaluation depends on the information and control functions provided by the interface. RoboPro~\citep{xie2025robopro} generates policy code from visual input, DAHLIA~\citep{meng2025dahlia} uses structured RGB-D feedback for replanning, and Reliable Code-as-Policies~\citep{ahn2025reliablecap} incorporates symbolic verification. VLCP~\citep{naouali2026vlcp} provides task-relevant simulator object poses, yet still observes grasp and placement failures. OpenETA~\citep{chen2026openeta} uses depth data to map selected pixels to 3D surface points, and in its full configuration, provides specialist perception and grasp modules~\citep{carion2025sam3,deitke2024molmo,sundermeyer2021contactgraspnet}. CaP-X~\citep{fu2026capx} evaluates tiers with different combinations of simulator-state access, specialist perception, action abstraction, examples, and visual feedback, making their individual effects difficult to isolate. \systemname fixes the robot API and assistance restrictions across configurations, leaving task-relevant 3D estimation, target selection, and policy construction to the agent. Table~\ref{tab:related} summarizes these interface differences.

\begin{table}[tb]
  \centering
  \caption{\textbf{Interface boundaries in Code-as-Policy manipulation.} The columns compare depth sensing, privileged object states, specialist perception or grasp outputs, and agent construction of task-relevant 3D estimates and grasp targets. CaP-X uses S for single-turn and M for multi-turn settings.
  }
  \label{tab:related}

  \setlength{\tabcolsep}{4pt}
  \begin{adjustbox}{max width=\linewidth}
\begin{tabular}{@{}lcccc@{}}
    \toprule
    Method
    & \shortstack{No depth\\sensing}
    & \shortstack{No privileged\\scene state}
    & \shortstack{No specialist\\outputs}
    & \shortstack{Agent-constructed\\3D \& grasp} \\
    \midrule

    Code as Policies~\citep{liang2023codeaspolicies}
    & \xmark & \cmark & \xmark & \xmark \\

    CaP-X S1~\citep{fu2026capx}
    & \xmark\rlap{\textsuperscript{$\dagger$}} & \xmark & \cmark & \xmark \\

    CaP-X S2--S4
    & \xmark & \cmark & \xmark & \xmark \\

    CaP-X M1--M4
    & \xmark & \cmark & \xmark & \xmark \\

    VLCP~\citep{naouali2026vlcp}
    & \cmark & \xmark & \cmark & \xmark \\

    OpenETA (full)~\citep{chen2026openeta}
    & \xmark & \cmark & \xmark & \xmark \\

    OpenETA (Codex plugin)
    & \xmark & \cmark & \cmark & \xmark\rlap{\textsuperscript{$\ddagger$}} \\

    Agentic Code-as-Policy (ours)
    & \cmark & \cmark & \cmark & \cmark \\
    \bottomrule
  \end{tabular}
\end{adjustbox}

  \par\smallskip
  \begin{minipage}{\linewidth}
    \small
    $^{\dagger}$CaP-X S1's LIBERO interface exposes simulated camera depth.
    
    \smallskip
    $^{\ddagger}$OpenETA's Codex plugin uses depth data to convert agent-selected image points into 3D coordinates.
  \end{minipage}
\end{table}

\noindent\textbf{Manipulation benchmarks and evaluation interfaces.}
RLBench, CALVIN, ManiSkill2, and LIBERO \citep{james2020rlbench,mees2022calvin,gu2023maniskill2,liu2023libero} provide tasks for evaluating manipulation policies. BEHAVIOR-1K~\citep{li2023behavior1k} and RoboCasa~\citep{nasiriany2024robocasa} broaden household task coverage, while BiGym~\citep{chernyadev2025bigym} targets mobile bimanual manipulation. SimplerEnv~\citep{li2024simplerenv} evaluates real-robot policies in simulation, and VLABench~\citep{zhang2025vlabench} focuses on long-horizon language-conditioned tasks. EmbodiedBench~\citep{yang2025embodiedbench} and RoboBench~\citep{luo2026robobench} evaluate multimodal language models as embodied agents. In contrast, \systemname focuses on how agents construct and revise executable policies during interaction. Built on RoboTwin~2.0~\citep{chen2025robotwin2}, it fixes the robot API, resource budgets, and hidden verifier across agent configurations, and records execution traces to analyze policy construction, physical progress, and failure.

\section{CodeActionBench}
\label{sec:benchmark}

\subsection{Tasks and evaluation settings}

\systemname comprises 25 manipulation tasks in SAPIEN~\citep{xiang2020sapien}, using RoboTwin~2.0's ALOHA-AgileX dual-arm robot configuration~\citep{chen2025robotwin2}. The tasks cover single-arm and bimanual grasping, transport, handover, tool use, arrangement, stacking, transient contact events, and manipulation of articulated objects. For consistent scene initialization, domain randomization is disabled and all configurations use the same fixed scene seed for each task.

Each attempt is a complete episode from reset to termination under a fixed agent configuration and resource budget. Task instructions state goals without disclosing verifier thresholds. Evaluation uses no task-specific fine-tuning, robot demonstrations, scripted action sequences, task planners, or pre-built skills. Table~\ref{tab:tasks} lists all tasks, budgets, and scene seeds.

\subsection{Agentic Code-as-Policy interface and information boundary}
\label{sec:interface}

The interface provides a shared robot API for observation, geometric computation, robot feedback, and bounded motion. Head and wrist cameras provide multi-view RGB observations, while calibrated geometry tools compute 3D estimates from agent-selected image points and geometric assumptions. Motion tools execute agent-specified Tool Center Point (TCP) poses and gripper commands, rejecting commands when planning fails and stopping motions that persistently stall. Agents may invoke tools directly or compose them into Python programs through \runcode, using calculations, branches, loops, and action sequences. Returned observations, robot state, contact feedback, and motion outcomes support policy revision across model turns. The agent decides how to organize execution and when to terminate through \texttt{done}, which records its completion judgment. Appendix~\ref{app:api} provides the detailed API documentation.

Both direct calls and programs obey the same information restrictions. The API does not expose exact object poses, task-specific grasp or placement targets, specialist perception outputs, or verifier state. Geometry tools perform the requested computations but do not select visual correspondences or verify the agent's geometric assumptions. The agent remains responsible for selecting evidence, interpreting estimates, and constructing manipulation targets. Task success is determined independently by the verifier, whose result is never returned to the agent.

\subsection{Reference harness}
\label{sec:harness}

The reference harness manages model interaction with the benchmark throughout each episode. It forms part of the evaluated agent configuration, while other agent systems may use their own harnesses with the same robot API. At each turn, it constructs the model request, converts provider-specific responses into a common format, validates tool calls against the API schemas, and records the interaction.

The harness follows a fixed context-management policy. Text interactions are appended to preserve a stable request prefix, while images are managed separately. Each request includes images from the two most recent observation rounds, with at most six images from any single \runcode result. Persistent observation identifiers and text records allow earlier images to be retrieved after removal from the current context. Near the context limit, the harness retains the eight most recent assistant messages and tool results verbatim and converts older interactions into at most 128 records using fixed rules. If the request still exceeds the limit, the episode terminates. Appendix~\ref{app:harness} provides further implementation details.

\subsection{Out-of-band task verification}
\label{sec:verification}

A hidden verifier assigns each attempt a binary outcome after termination, without returning the result to the agent. For 21 tasks, success is determined by the final physical state. The remaining four tasks are scored by whether a required event occurs during execution. Table~\ref{tab:tasks} specifies the scoring rule for each task. An episode ends when the agent calls \texttt{done}, exhausts a budget, or meets another predefined termination condition. The \texttt{done} call records the agent's completion claim without providing success feedback, allowing this claim to be compared with the independent verifier outcome.

\subsection{Protocol and metrics}
\label{sec:protocol}

We evaluate nine agent configurations, each defined by its foundation model and harness. Seven share the reference harness, while two use Claude Code~\citep{anthropicClaudeCode} and Codex CLI~\citep{openaiCodexCLI} as harnesses,  respectively. Comparisons involving different harnesses therefore evaluate complete agent systems. All configurations use the same environment, robot API, task-specific budgets, and verifier. Each task uses one fixed scene seed, with three valid attempts per configuration, yielding 75 attempts per configuration and 675 overall. Superseded runs and documented infrastructure failures are excluded.

We report success rate over the 75 attempts and task coverage, defined as the number of tasks solved at least once in three attempts. Coverage measures performance on the fixed task instances rather than generalization across scenes. End-to-end execution time includes model-provider waiting and environment execution. Process analyses use recorded tool calls, programs, robot actions, and scene states, with the eligible sample reported for each measurement. These diagnostics do not affect the success-based ranking.

\section{Evaluations}
\label{sec:results}

We evaluate nine agent configurations on 25 fixed task instances, with three attempts per instance for each configuration. Across all configurations, 184 of 675 attempts succeed (27.3\%). We report task performance and inference cost, followed by analyses of policy execution and task completion assessment.

\subsection{Performance across configurations and tasks}

\begin{figure}[!t]
  \centering
  \includegraphics[width=\linewidth]{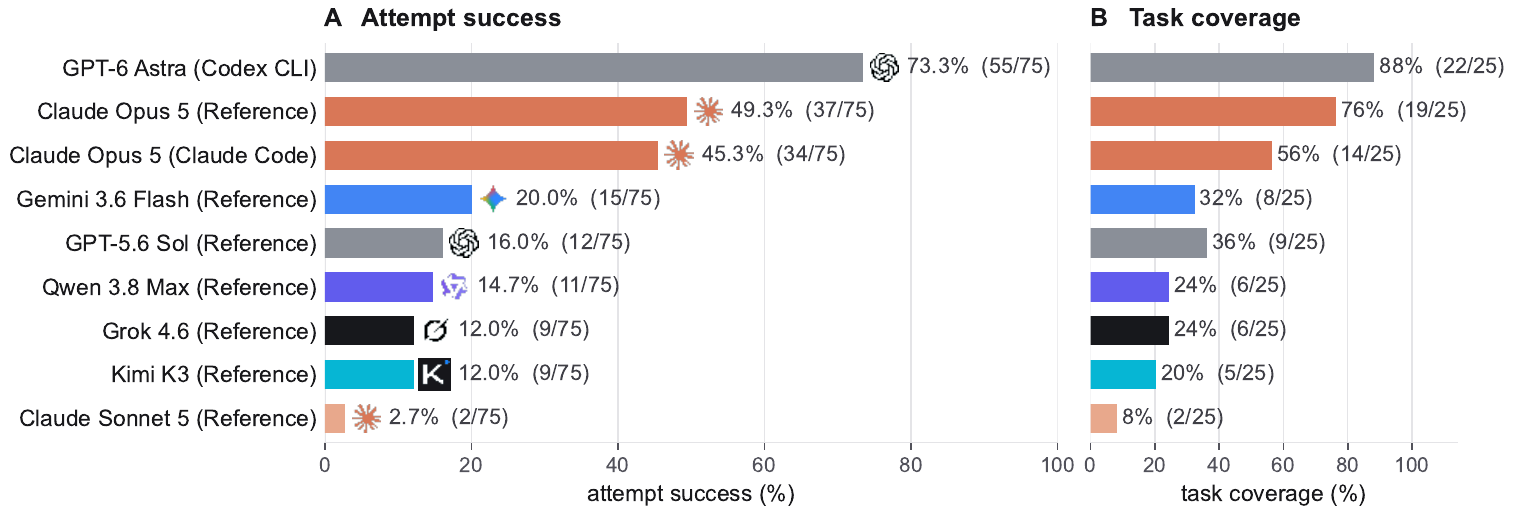}
  \caption{\textbf{Task performance.} (A)~Success rate over 75 attempts. (B)~Percentage of the 25 fixed instances solved at least once in three independent attempts.}
  \label{fig:outcomes}
\end{figure}

As shown in Figure~\ref{fig:outcomes}, GPT-6 Astra (Codex CLI) achieves the highest success rate of 73.3\% and solves 22/25 tasks. Claude Opus~5 with the reference harness follows with 49.3\% success and 19/25 tasks solved, compared with 45.3\% and 14/25 under Claude Code. The two Opus configurations thus differ by only 4.0 percentage points in success rate but by five tasks in coverage.

Among the seven configurations using the reference harness, Claude Opus~5 leads in both metrics. It achieves a success rate of 49.3\%, followed by Gemini~3.6 Flash at 20.0\%. Its task coverage reaches 76\%, followed by GPT-5.6 Sol at 36\%. Table~\ref{tab:consistency} reports success counts, task coverage, and consistency across the three attempts.

\subsection{Cost and time efficiency}
\label{sec:cost}

\begin{figure}[!t]
  \centering
  \includegraphics[width=\linewidth]{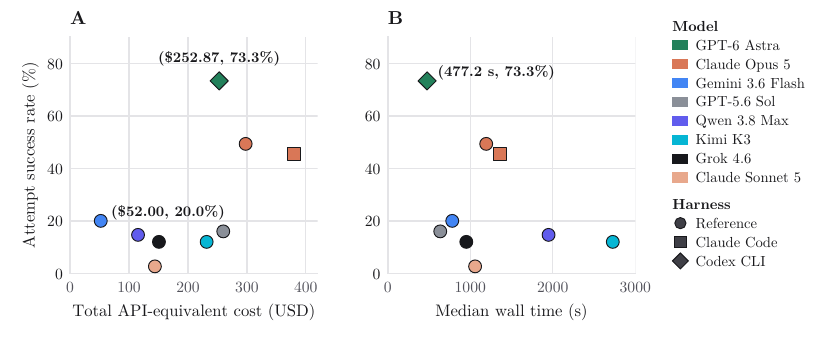}
  \caption{\textbf{Success rate against inference cost and wall time.} Each point represents one configuration's 75 attempts. Both panels show success rate vertically; the horizontal axes are (A)~total API-equivalent inference cost and (B)~median wall time per attempt.}
  \label{fig:cost-time}
\end{figure}

Figure~\ref{fig:cost-time} compares success rates with total API-equivalent inference cost over 75 attempts and median end-to-end execution time. Astra achieves higher success at lower cost and with shorter execution time than either Opus configuration. Compared with Opus under the reference harness, Astra achieves 73.3\% versus 49.3\% success, with 15.1\% lower cost and 60.0\% shorter median execution time. Its median of 477.2\,s is the shortest among all nine configurations. Gemini~3.6 Flash has the lowest total cost of \$52.00 and a success rate of 20.0\%, outperforming five other reference-harness configurations at lower cost. Appendix~\ref{app:cost} provides the pricing assumptions and resource distributions.

\subsection{Policy execution analysis}
\label{sec:construction}

Binary task outcomes do not distinguish an initial grasp failure from a placement failure after successful transport. We therefore use \emph{subgoal checkpoints} to record completion of intermediate task requirements. We also use \emph{spatial checkpoints} to assess whether the robot reaches locations associated with these subgoals. We compare coverage of the two checkpoint types, and plot cumulative spatial coverage against charged tool calls for all 75 attempts per configuration, using the same 25 task instances across nine configurations. We then use spatial checkpoint matches and action records to identify which subgoals were not completed and where execution failed.

\noindent\textbf{Checkpoint definitions.} We define 55 subgoal checkpoints and 72 spatial checkpoints across the 25 tasks, using the same criteria for all configurations. Subgoal criteria are derived from task verifiers and specify object states, spatial relations, or required interaction events. Spatial checkpoints define regions for the robot's tool center point (TCP), with reference positions and tolerances determined from task geometry and successful expert executions~\citep{chen2025robotwin2}. Regions associated with movable targets follow the corresponding objects. Subgoal completion is assessed from recorded scene and robot states, while spatial matches use recorded end-effector positions or measured position bounds. Each checkpoint is counted once, at its first confirmed completion or match. Spatial matches must satisfy any prerequisite checkpoint dependencies, which apply only to trajectory analysis and do not constrain the agent during execution.

For each attempt, coverage is the fraction of its subgoal or spatial checkpoints attained. We average the three attempts within each task and then average across the 25 tasks with equal weight. At charged call $k$, cumulative spatial coverage includes matches recorded within the first $k$ calls. After an attempt ends, its coverage remains unchanged at subsequent call counts. Repeated matches do not increase coverage, and later regression does not remove an earlier match. Observation, computation, and unsuccessful operations all count toward the call budget, while one program submission may match several checkpoints. Checkpoints without confirmed matches remain in the denominator, so both measures depend on recording completeness. Appendices~\ref{app:policy-checkpoints} and~\ref{app:checkpoint-aggregation} provide the definitions, calibration details, and evidence-recovery procedures.

\begin{figure}[!t]
  \centering
  \includegraphics[width=\linewidth]{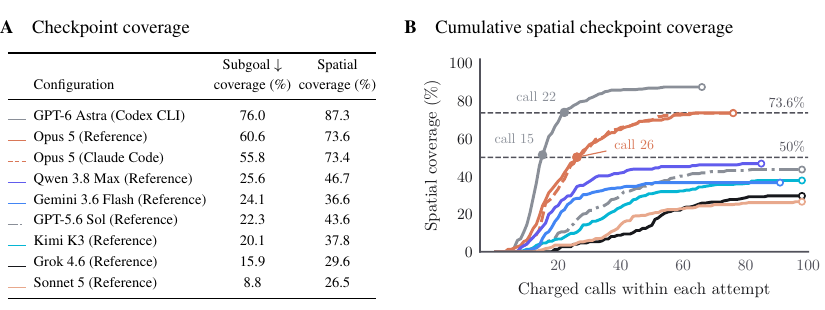}
  \caption{\textbf{Subgoal and spatial checkpoint coverage across configurations.} Panel A reports final coverage in descending order of subgoal checkpoint coverage. Panel B shows spatial checkpoint coverage over charged calls within each attempt. Each measure is averaged over three attempts per task, and all 25 tasks receive equal weight.}
  \label{fig:policy-progress}
\end{figure}

\noindent\textbf{Spatial coverage over tool calls.} Astra leads both checkpoint types in coverage (Figure~\ref{fig:policy-progress}). Only three configurations exceed 50\% spatial coverage: Astra at 87.3\%, Opus (Reference) at 73.6\%, and Opus (Claude Code) at 73.4\%. By call 22, Astra exceeds every other configuration's final spatial coverage. The two Opus configurations achieve nearly identical spatial coverage despite differences in attempt success rate and task coverage.

\begin{figure}[!t]
  \centering
  \includegraphics[width=\linewidth]{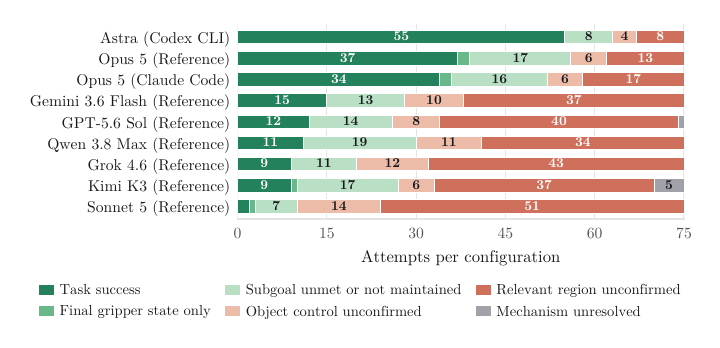}
  \caption{\textbf{Task outcomes and failure stages.} The horizontal axis counts attempts, from 0 to 75 per configuration. Configurations are ordered by task success rate. Each attempt contributes once to success or the earliest supported failure stage; unresolved cases remain separate.}
  \label{fig:failure-stages}
\end{figure}
\noindent\textbf{Failure analysis.} Each failed attempt is assigned to its earliest evidence-supported failure stage, using unmet or lost subgoals, associated spatial checkpoints, and action records of object motion, contact, and release. Figure~\ref{fig:failure-stages} groups attempts according to whether arrival at the relevant region and object control are confirmed, and whether the subgoal is completed and maintained. Failures limited to the final gripper condition and cases with insufficient evidence are reported separately. Appendix~\ref{app:failure-analysis} provides the assignment criteria.

Among the 491 failed attempts, 280 have no confirmed arrival at the relevant region, 77 have no confirmed object control, and 122 reach the operation region but do not complete the subgoal or later lose it. Six fail only the final gripper condition, and six remain unresolved. Astra records 20 failures, compared with 38 for Opus Reference and 41 for Opus Claude Code. Arrival is unconfirmed in 8, 13, and 17 attempts, respectively, compared with 34 to 51 for each of the other six configurations. This result indicates that stronger systems have fewer failures with unconfirmed arrival at the relevant operation region. Object-control and subgoal-outcome failures together account for more than half of the failures in these three configurations (12/20, 23/38, and 22/41). These later-stage cases are consistent with difficulties in object retention, precise contact or placement, and state maintenance.

\subsection{Task completion assessment and termination}
\label{sec:completion-judgment}

Agents should assess task completion from RGB observations, robot state, contact feedback, and execution outcomes, without access to the verifier's verdict. They are instructed to call \texttt{done(report=..., success\_\allowbreak{}claim=...)} when the task is complete or they cannot proceed. This call ends the episode and records the agent's account of its actions and completion judgment. We compare the Boolean claim with the independent verifier outcome. An \emph{overclaim} reports success on a failed attempt, while an \emph{underclaim} reports failure on a successful attempt. Attempts without an explicit claim are recorded separately rather than treated as incorrect judgments.

\begin{table}[tb]
  \centering
  \caption{\textbf{Termination and completion judgment.}
  A uses all 75 attempts per configuration.
  B and C partition successful and failed attempts,
  respectively, under corrected verifier outcomes.
  Overclaim: a success claim on a verifier-failed attempt. Underclaim: a failure claim on a verifier-successful attempt. Attempts without an explicit claim are reported separately.}
  \label{tab:completion-judgment}

  \small
  \setlength{\tabcolsep}{4pt}
  \renewcommand{\arraystretch}{1.0}
  \begin{adjustbox}{max width=\linewidth}
\begin{tabular}{@{}lrrr@{}}
    \toprule
    Model
    & \multicolumn{1}{c}{GPT-6 Astra}
    & \multicolumn{2}{c}{Claude Opus 5} \\
    \cmidrule(lr){2-2}
    \cmidrule(lr){3-4}
    Harness
    & \multicolumn{1}{c}{Codex CLI}
    & \multicolumn{1}{c}{Reference harness}
    & \multicolumn{1}{c}{Claude Code} \\
    \midrule

    \multicolumn{4}{l}{\textbf{A. All attempts}} \\
    Ends with \texttt{done}
    & 72/75 (96.0\%)
    & 63/75 (84.0\%)
    & 54/75 (72.0\%) \\

    Ends with a correct claim
    & 68/75 (90.7\%)
    & 50/75 (66.7\%)
    & 42/75 (56.0\%) \\

    \addlinespace[2pt]
    \multicolumn{4}{l}{\textbf{B. Successful attempts}} \\
    Correct success claim
    & 54/55 (98.2\%)
    & 36/37 (97.3\%)
    & 30/34 (88.2\%) \\

    Underclaim
    & 1/55 (1.8\%)
    & 0/37 (0.0\%)
    & 1/34 (2.9\%) \\

    No explicit claim
    & 0/55 (0.0\%)
    & 1/37 (2.7\%)
    & 3/34 (8.8\%) \\

    \addlinespace[2pt]
    \multicolumn{4}{l}{\textbf{C. Failed attempts}} \\
    Correct failure claim
    & 14/20 (70.0\%)
    & 14/38 (36.8\%)
    & 12/41 (29.3\%) \\

    Overclaim
    & 3/20 (15.0\%)
    & 13/38 (34.2\%)
    & 11/41 (26.8\%) \\

    No explicit claim
    & 3/20 (15.0\%)
    & 11/38 (28.9\%)
    & 18/41 (43.9\%) \\
    \bottomrule
  \end{tabular}
\end{adjustbox}
\end{table}

Table~\ref{tab:completion-judgment} reports how often agents terminate through \texttt{done} and how often they terminate with a correct claim, both measured over all 75 attempts. Correct claims include both success and failure reports that agree with the verifier. Astra calls \texttt{done} in 72/75 attempts, compared with 63/75 for Opus Reference and 54/75 for Opus Claude Code. It provides a correct claim in 68/75 attempts (90.7\%), compared with 50/75 (66.7\%) and 42/75 (56.0\%), respectively. These proportions reflect both whether the agent supplies a claim before termination and whether that claim is correct.

Separating successful and failed attempts reveals where the configurations differ. Among successful attempts, Astra, Opus Reference, and Opus Claude Code correctly report success in 98.2\%, 97.3\%, and 88.2\% of cases, respectively. Among failed attempts, Astra correctly reports failure in 70.0\% of cases, compared with 36.8\% and 29.3\% for the Opus systems. Incorrect success claims account for 15.0\%, 34.2\%, and 26.8\% of their failed attempts, respectively. Underclaims occur once for Astra and once for Opus Claude Code. Astra thus more often ends with a correct assessment, including when manipulation has not achieved the task goal.

Budget limits can end an episode before the agent reports its assessment. All attempts without an explicit claim in these three configurations terminate at a tool-call or physical-time limit, including one successful Opus Reference attempt and three successful Opus Claude Code attempts. This pattern is more pronounced for Claude Sonnet~5, which leaves 70/75 attempts without an explicit claim. Of these, 69 end at a budget limit and one ends through \texttt{done} without a Boolean claim. Missing claims therefore need to be distinguished from incorrect assessments, since forced termination leaves the agent's final judgment unobserved. Appendix~\ref{app:termination} reports all configurations and stopping conditions.

\section{Limitations}
\label{sec:limitations}

\noindent\textbf{Evaluation scope.} We evaluate 25 simulated manipulation tasks with a single dual-arm embodiment and one fixed scene instance per task. Repeated attempts measure variation in agent behavior on the same instances. Generalization across scenes, task families, and embodiments requires further evaluation. Transfer to physical robots also remains untested.

\noindent\textbf{Evaluation cost and latency.} Inference cost and wall time limit the number of configurations and attempts in the evaluation. Sequential model calls also delay action selection and policy revision. This latency constrains time-sensitive manipulation.

\section{Conclusion}
\label{sec:conclusion}

We introduce \systemname, a benchmark for evaluating agentic Code-as-Policy in embodied manipulation. Agents construct and revise executable policies from visual observations and execution feedback through a shared robot API, without access to privileged scene state or specialist perception and grasp outputs. Across nine agent configurations on 25 fixed simulated task instances, success rates range from 2.7\% to 73.3\% without additional task-specific fine-tuning or demonstrations. Trajectory analysis shows that failed attempts can reach intermediate spatial checkpoints, while these spatial matches alone do not establish successful manipulation. By combining task outcomes, execution traces, and agent-reported completion claims, \systemname supports evaluation of policy execution and task completion assessment under a common protocol.

\bibliographystyle{codeactionbench}
\bibliography{references}
\appendix
\raggedbottom

\section{Evaluation Protocol and Task Suite}
\label{app:protocol}

\subsection{Environment and fixed task instances}
\label{app:tasks}
The evaluation uses SAPIEN~\citep{xiang2020sapien} and RoboTwin~2.0's ALOHA-AgileX dual-arm configuration~\citep{chen2025robotwin2}. We disable random perturbations to the background, clutter, lighting, head camera, and table height. For each task, all configurations use the same fixed scene seed. These choices hold the physical task instance constant without exposing scene-object state to the agent.

The head camera provides $640\times480$ RGB images and the two wrist cameras provide $320\times240$ RGB images. Each observation includes camera intrinsics and extrinsics, with wrist-camera extrinsics corresponding to the arm pose at capture time. The agent can use these measurements to infer geometry from RGB views, but receives neither depth images nor object poses. Appendix~\ref{app:api} specifies the available operations.

Each configuration makes three attempts on each of the 25 fixed task instances. Model sampling seeds are not fixed, so the three attempts measure variation on the same scene. Outcome statistics include attempts that end at a resource limit or a runtime error.

\subsection{Resource budgets and stopping rules}
\label{app:budgets}

Each task has a tool-call budget and a simulation-time budget, fixed before evaluation and shared across all configurations. The tool-call budget limits API calls and program submissions, while the simulation-time budget limits cumulative physical execution, including actions performed within programs.

For each task, let $t_{\mathrm{expert,min}}$ denote the estimated expert completion time in minutes and $t_{\mathrm{expert,sim}}$ the simulated duration of an official demonstration in seconds. We set
\[
B_{\mathrm{tool}}=\left\lceil14t_{\mathrm{expert,min}}\right\rceil,\qquad
B_{\mathrm{sim}}=\operatorname{ceil}_{5\mathrm{s}}
\left(\max(60\mathrm{s},5t_{\mathrm{expert,sim}})\right),
\]
where $\operatorname{ceil}_{5\mathrm{s}}$ rounds upward to the nearest multiple of five seconds. Table~\ref{tab:tasks} reports the resulting budgets for each task.

Each direct API call or \runcode submission counts as one tool call, regardless of execution success. A program can therefore combine multiple API operations into a single tool call, while all robot actions it executes remain subject to the simulation-time budget. This budget measures physical execution in simulated seconds and excludes model inference and provider waiting.

An attempt ends when the agent calls \texttt{done}, exhausts a resource budget, or encounters an error that prevents further interaction. The verifier evaluates the task outcome at termination, including attempts stopped by budget exhaustion.

\subsection{Alignment of task instructions and verification}
\label{app:verification}

Task verification builds on RoboTwin~2.0's success checks. Evaluating agents without task demonstrations or task-specific fine-tuning requires clear agreement between instructions and scoring criteria. The benchmark therefore adapts selected instructions and checks to make the required outcomes explicit and accept valid solutions to the stated tasks.

For requirements with a clear natural-language description, the instructions state the scoring conditions directly. Examples include keeping the pot level during lifting and holding the small bin above the tabletop after pouring. Adding these requirements clarifies the intended outcome while leaving perception, target selection, and motion planning to the agent.

Some original checks also enforce choices from the expert policy. For laptop opening, RoboTwin selects an arm according to the laptop's initial orientation and checks that arm's proximity to the lid. The instruction leaves arm choice open, so the adapted verifier removes this assignment while preserving the opening-angle and lid-proximity conditions. Similarly, RoboTwin assigns each shoe to a specific target in the two-shoe placement task, despite the instruction leaving this assignment unspecified. Accepting either assignment preserves the position, orientation, and gripper-opening requirements without prescribing shoe identity. Encoding these choices through descriptions of the initial scene could also introduce ambiguity under future scene randomization.

RoboTwin's expert demonstrations also guide the roller-lifting and bread-placement adaptations. For roller lifting, the instruction specifies a bimanual grasp, and the verifier adds contact checks for both grippers alongside the original gripper-closure and lifting conditions. For bread placement, the instruction specifies lifting the skillet with one arm and placing the bread into it with the other. The verifier then checks the bread's position relative to the skillet and its release from both grippers. Replacing the original absolute-height checks prevents accepting bread still held above the skillet. The demonstrations guide benchmark construction but remain unavailable to evaluated agents.

RoboTwin checks success during execution and records a successful outcome once the task condition holds. Here, agents can continue acting after reaching that state. Scoring 21 tasks at termination therefore requires agents to preserve the intended outcome through the end of the attempt. The remaining four tasks---hammer striking, bell pressing, stapler pressing, and bin pouring---record the required events during execution, preserving RoboTwin's treatment of event completion. Their verifiers also check any additional final-state requirements at termination. Table~\ref{tab:tasks} specifies each task's scoring rule. Verifier outcomes remain hidden from the agent throughout the interaction.

\subsection{Reference solutions and budget feasibility}
\label{app:oracle}

To verify that the aligned instructions and scoring rules define achievable tasks under the available interface and budgets, task-specific reference solutions are constructed for all 25 fixed task instances. These solutions are developed using privileged scene information and execute robot actions through the same public API available to evaluated agents. All 25 instances are successfully completed within their tool-call and simulation-time budgets. Table~\ref{tab:tasks} reports the API-call counts and simulated execution times of these solutions. The reference scripts and privileged scene information are withheld from evaluated agents.

\subsection{Data availability and reproducibility}
We plan to publicly release the benchmark code, evaluation configurations, agent interaction logs, and analysis scripts to enable verification of the reported results and further evaluation under the same protocol. Section~\ref{sec:benchmark} and Appendices~\ref{app:protocol}--\ref{app:harness} describe the task instances, scene seeds, robot interface, agent configurations, and evaluation rules. Appendices~\ref{app:construction} and~\ref{app:failure-analysis} document checkpoint analysis, replay validation, and failure-stage assignment. Two factors limit exact reproducibility. First, the evaluated models are accessed through provider APIs or vendor harnesses; model updates, retirement, and changes to these services may prevent identical reruns. Second, stochastic model outputs and variation in motion planning can lead to different action sequences and physical outcomes. We evaluate each configuration with three attempts on each of 25 fixed task instances, yielding 675 attempts across nine configurations. These repetitions capture variation on the evaluated instances, although additional attempts would provide more precise estimates.

\subsection{Task catalog}
\label{app:task-catalog}
Table~\ref{tab:tasks} lists the task goals, fixed scene seeds, scoring modes, and resource budgets. Reference-solution times are rounded to the nearest second.

\begingroup
\footnotesize
\setlength{\tabcolsep}{3pt}
\renewcommand{\arraystretch}{1.0}
\setlength{\LTcapwidth}{\linewidth}
\begin{longtable}{@{}>{\raggedright\arraybackslash}p{.30\linewidth}*{6}{>{\raggedleft\arraybackslash}p{\dimexpr(.70\linewidth-12\tabcolsep)/6\relax}}@{}}
\caption{\textbf{Tasks and evaluation budgets.} Each of the 25 fixed task instances is listed with a representative instruction, scene seed, scoring rule, and reference-solution resource use. $B_{\mathrm{tool}}$ and $B_{\mathrm{sim}}$ bound charged calls and simulated seconds. Scoring uses required events during execution (latch) or physical conditions at termination (final).}
\label{tab:tasks}\label{tab:tasks-b}\\
\toprule
Task & Seed & Scoring & $B_{\mathrm{tool}}$ & \shortstack{$B_{\mathrm{sim}}$\\(s)} & \shortstack{Solution\\API calls} & \shortstack{Solution\\sim (s)} \\
\midrule
\endfirsthead
\multicolumn{7}{l}{\textit{Table~\thetable\ continued.}}\\
\toprule
Task & Seed & Scoring & $B_{\mathrm{tool}}$ & \shortstack{$B_{\mathrm{sim}}$\\(s)} & \shortstack{Solution\\API calls} & \shortstack{Solution\\sim (s)} \\
\midrule
\endhead
\midrule
\multicolumn{7}{r}{\textit{Continued on the next page.}}\\
\endfoot
\bottomrule
\endlastfoot
\multicolumn{7}{l}{\emph{Tool use and contact processes (4 tasks)}} \\
\addlinespace[1pt]
\texttt{beat\_\allowbreak{}block\_\allowbreak{}hammer} & 0 & latch & 42 & 60 & 40 & 28 \\*
\multicolumn{7}{@{\hspace{1.2em}}>{\raggedright\arraybackslash}p{\dimexpr\linewidth-2\tabcolsep-1.2em}}{{There is a hammer and a block on the table, use the arm to grab the hammer and beat the block.}} \\
\addlinespace[3pt]
\texttt{click\_\allowbreak{}bell} & 0 & latch & 28 & 60 & 20 & 23 \\*
\multicolumn{7}{@{\hspace{1.2em}}>{\raggedright\arraybackslash}p{\dimexpr\linewidth-2\tabcolsep-1.2em}}{{Click the bell's top center on the table. Use the arm on the bell's side and keep its gripper closed.}} \\
\addlinespace[3pt]
\texttt{press\_\allowbreak{}stapler} & 0 & latch & 28 & 60 & 20 & 23 \\*
\multicolumn{7}{@{\hspace{1.2em}}>{\raggedright\arraybackslash}p{\dimexpr\linewidth-2\tabcolsep-1.2em}}{{Use one arm to press the stapler.}} \\
\addlinespace[3pt]
\texttt{scan\_\allowbreak{}object} & 4 & final & 70 & 60 & 52 & 30 \\*
\multicolumn{7}{@{\hspace{1.2em}}>{\raggedright\arraybackslash}p{\dimexpr\linewidth-2\tabcolsep-1.2em}}{{Use one arm to pick the scanner and use the other arm to pick the object, and use the scanner to scan the object. Keep the scanner head close and directly aligned with the object.}} \\
\addlinespace[3pt]
\cmidrule(lr){1-7}
\multicolumn{7}{l}{\emph{Multi-object organization and assembly (11 tasks)}} \\
\addlinespace[1pt]
\texttt{blocks\_\allowbreak{}ranking\_\allowbreak{}size} & 0 & final & 84 & 125 & 73 & 54 \\*
\multicolumn{7}{@{\hspace{1.2em}}>{\raggedright\arraybackslash}p{\dimexpr\linewidth-2\tabcolsep-1.2em}}{{There are three blocks on the table, the color of the blocks is random, move the blocks to the center of the table, and arrange them from largest to smallest, from left to right. Keep the row compact.}} \\
\addlinespace[3pt]
\texttt{hanging\_\allowbreak{}mug} & 0 & final & 70 & 90 & 57 & 43 \\*
\multicolumn{7}{@{\hspace{1.2em}}>{\raggedright\arraybackslash}p{\dimexpr\linewidth-2\tabcolsep-1.2em}}{{Use left arm to pick the mug on the table, rotate the mug and put the mug down in the middle of the table, use the right arm to pick the mug and hang it onto the rack.}} \\
\addlinespace[3pt]
\texttt{place\_\allowbreak{}bread\_\allowbreak{}basket} & 2 & final & 56 & 85 & 55 & 31 \\*
\multicolumn{7}{@{\hspace{1.2em}}>{\raggedright\arraybackslash}p{\dimexpr\linewidth-2\tabcolsep-1.2em}}{{Put every piece of bread on the table into the basket.}} \\
\addlinespace[3pt]
\texttt{place\_\allowbreak{}bread\_\allowbreak{}skillet} & 5 & final & 56 & 60 & 56 & 43 \\*
\multicolumn{7}{@{\hspace{1.2em}}>{\raggedright\arraybackslash}p{\dimexpr\linewidth-2\tabcolsep-1.2em}}{{If there is one bread on the table, use one arm to lift the skillet and the other arm to grab the bread and put it into the skillet.}} \\
\addlinespace[3pt]
\texttt{place\_\allowbreak{}cans\_\allowbreak{}plasticbox} & 0 & final & 56 & 80 & 55 & 43 \\*
\multicolumn{7}{@{\hspace{1.2em}}>{\raggedright\arraybackslash}p{\dimexpr\linewidth-2\tabcolsep-1.2em}}{{Use dual arm to pick and place cans into plasticbox.}} \\
\addlinespace[3pt]
\texttt{place\_\allowbreak{}dual\_\allowbreak{}shoes} & 6 & final & 98 & 90 & 73 & 55 \\*
\multicolumn{7}{@{\hspace{1.2em}}>{\raggedright\arraybackslash}p{\dimexpr\linewidth-2\tabcolsep-1.2em}}{{Use both arms to pick up the two shoes on the table and put them in the shoebox, with the shoe tip pointing to the left.}} \\
\addlinespace[3pt]
\texttt{place\_\allowbreak{}mouse\_\allowbreak{}pad} & 0 & final & 56 & 60 & 41 & 24 \\*
\multicolumn{7}{@{\hspace{1.2em}}>{\raggedright\arraybackslash}p{\dimexpr\linewidth-2\tabcolsep-1.2em}}{{Grab the mouse and place it on a colored mat. Center and align the mouse on the mat.}} \\
\addlinespace[3pt]
\texttt{place\_\allowbreak{}object\_\allowbreak{}basket} & 0 & final & 84 & 70 & 63 & 39 \\*
\multicolumn{7}{@{\hspace{1.2em}}>{\raggedright\arraybackslash}p{\dimexpr\linewidth-2\tabcolsep-1.2em}}{{Put the object in the basket, then pick the basket up.}} \\
\addlinespace[3pt]
\texttt{put\_\allowbreak{}bottles\_\allowbreak{}dustbin} & 0 & final & 84 & 175 & 82 & 64 \\*
\multicolumn{7}{@{\hspace{1.2em}}>{\raggedright\arraybackslash}p{\dimexpr\linewidth-2\tabcolsep-1.2em}}{{Use arms to grab the bottles and put them into the dustbin to the left of the table.}} \\
\addlinespace[3pt]
\texttt{stack\_\allowbreak{}blocks\_\allowbreak{}three} & 0 & final & 70 & 130 & 66 & 53 \\*
\multicolumn{7}{@{\hspace{1.2em}}>{\raggedright\arraybackslash}p{\dimexpr\linewidth-2\tabcolsep-1.2em}}{{There are three blocks on the table, the color of the blocks is red, green and blue, move the blocks to the center of the table, and stack the blue block on the green block, and the green block on the red block.}} \\
\addlinespace[3pt]
\texttt{stack\_\allowbreak{}bowls\_\allowbreak{}three} & 0 & final & 84 & 135 & 81 & 55 \\*
\multicolumn{7}{@{\hspace{1.2em}}>{\raggedright\arraybackslash}p{\dimexpr\linewidth-2\tabcolsep-1.2em}}{{Stack the three bowls on top of each other.}} \\
\addlinespace[3pt]
\cmidrule(lr){1-7}
\multicolumn{7}{l}{\emph{Dual-arm cooperative manipulation (4 tasks)}} \\
\addlinespace[1pt]
\texttt{dump\_\allowbreak{}bin\_\allowbreak{}bigbin} & 3 & latch & 84 & 125 & 66 & 33 \\*
\multicolumn{7}{@{\hspace{1.2em}}>{\raggedright\arraybackslash}p{\dimexpr\linewidth-2\tabcolsep-1.2em}}{{Grab the small bin and pour the balls into the big bin. Finish with the small bin held above the tabletop.}} \\
\addlinespace[3pt]
\texttt{grab\_\allowbreak{}roller\_\allowbreak{}dual\_\allowbreak{}contact} & 0 & final & 42 & 60 & 33 & 19 \\*
\multicolumn{7}{@{\hspace{1.2em}}>{\raggedright\arraybackslash}p{\dimexpr\linewidth-2\tabcolsep-1.2em}}{{Use both arms to grab and lift the roller on the table.}} \\
\addlinespace[3pt]
\texttt{lift\_\allowbreak{}pot} & 0 & final & 42 & 60 & 42 & 25 \\*
\multicolumn{7}{@{\hspace{1.2em}}>{\raggedright\arraybackslash}p{\dimexpr\linewidth-2\tabcolsep-1.2em}}{{Use both arms to lift the pot, keeping it level.}} \\
\addlinespace[3pt]
\texttt{pick\_\allowbreak{}diverse\_\allowbreak{}bottles} & 0 & final & 56 & 60 & 43 & 29 \\*
\multicolumn{7}{@{\hspace{1.2em}}>{\raggedright\arraybackslash}p{\dimexpr\linewidth-2\tabcolsep-1.2em}}{{Pick up both bottles together in front of the robot within the camera view, and hold them there.}} \\
\addlinespace[3pt]
\multicolumn{7}{l}{\emph{Handover, regrasp and reorientation (4 tasks)}} \\
\addlinespace[1pt]
\texttt{handover\_\allowbreak{}block} & 0 & final & 42 & 80 & 42 & 33 \\*
\multicolumn{7}{@{\hspace{1.2em}}>{\raggedright\arraybackslash}p{\dimexpr\linewidth-2\tabcolsep-1.2em}}{{Use the left arm to grasp the red block on the table, handover it to the right arm and place it on the blue pad.}} \\
\addlinespace[3pt]
\texttt{handover\_\allowbreak{}mic} & 0 & final & 56 & 60 & 40 & 28 \\*
\multicolumn{7}{@{\hspace{1.2em}}>{\raggedright\arraybackslash}p{\dimexpr\linewidth-2\tabcolsep-1.2em}}{{Use one arm to grasp the microphone on the table and handover it to the other arm. Finish with the receiving arm holding it raised on that side.}} \\
\addlinespace[3pt]
\texttt{move\_\allowbreak{}can\_\allowbreak{}pot} & 0 & final & 70 & 60 & 49 & 60 \\*
\multicolumn{7}{@{\hspace{1.2em}}>{\raggedright\arraybackslash}p{\dimexpr\linewidth-2\tabcolsep-1.2em}}{{Use one arm to pick up the can and set it down upright beside the pot, on the side where it started, aligned with the pot.}} \\
\addlinespace[3pt]
\texttt{rotate\_\allowbreak{}qrcode} & 0 & final & 56 & 60 & 32 & 17 \\*
\multicolumn{7}{@{\hspace{1.2em}}>{\raggedright\arraybackslash}p{\dimexpr\linewidth-2\tabcolsep-1.2em}}{{Use one arm to rotate the QR-code board face-up, then leave it flat on the table.}} \\
\addlinespace[3pt]
\cmidrule(lr){1-7}
\multicolumn{7}{l}{\emph{Articulated storage and devices (2 tasks)}} \\
\addlinespace[1pt]
\texttt{open\_\allowbreak{}laptop\_\allowbreak{}setup\_\allowbreak{}arm} & 0 & final & 56 & 60 & 34 & 13 \\*
\multicolumn{7}{@{\hspace{1.2em}}>{\raggedright\arraybackslash}p{\dimexpr\linewidth-2\tabcolsep-1.2em}}{{Use one arm to open the laptop and keep the acting gripper at the lid edge.}} \\
\addlinespace[3pt]
\texttt{open\_\allowbreak{}microwave} & 0 & final & 56 & 140 & 47 & 21 \\*
\multicolumn{7}{@{\hspace{1.2em}}>{\raggedright\arraybackslash}p{\dimexpr\linewidth-2\tabcolsep-1.2em}}{{Use one arm to open the microwave door wide.}} \\
\addlinespace[3pt]
\end{longtable}
\endgroup

\section{Robot Interface and Information Boundary}
\label{app:api}

\subsection{Information access}
\label{app:access}
\label{app:interaction-protocol}
The robot API supports observation, spatial estimation, target construction, and physical execution through either direct calls or agent-written programs. Agents use RGB images, camera calibration, robot state, and action feedback to construct manipulation policies. Simulator source, object states, task-specific targets, and verifier results remain inaccessible; image processing operates on the RGB observations available to the agent.

Robot motion and reachability checks share the cuRobo planner~\citep{sundaralingam2023curobo}, which computes joint trajectories for agent-specified TCP targets. Its collision model includes robot self-collision but excludes the table and task objects. Planning therefore tests motion feasibility under this restricted model, while the agent uses visual and contact feedback to reason about interactions with the scene.

\subsection{Conventions and result semantics}
\label{app:api-conventions}
The core interface comprises 29 robot tools, \runcode for program execution, and \texttt{done} for termination. Tables~\ref{tab:api-observation}, ~\ref{tab:api-geometry} and ~\ref{tab:api-motion} summarize the inputs and outputs of the robot tools; the following sections explain their roles in policy construction and execution.

Positions and displacements are expressed in metres in the world frame: positive $x$ points to the robot's right, positive $y$ forward toward the table, and positive $z$ upward. Orientations use $(w,x,y,z)$ quaternions. Image points use pixel coordinates $(u,v)$ and an observation identifier, which associates each point with the image and camera calibration used for geometric calculations.

Geometric results include the computed estimate, its numerical validity, and uncertainty or diagnostics where available. Their interpretation depends on the agent's selected pixels and geometric assumptions. Motion results report requested targets, measured robot state, and planning or execution outcomes, allowing the agent to assess the achieved motion and revise subsequent commands. Task success is evaluated separately by the hidden verifier.

\subsection{Robot state and visual observations}
The agent can query the current TCP pose and gripper opening of each arm. Gripper dimensions and camera calibration are also available for spatial calculations. The fixed head camera provides a workspace overview, while wrist cameras provide views that change with arm motion. Agents can request individual images or a bundle captured at the same simulation state, then annotate selected pixels with \texttt{draw\_\allowbreak{}marks} to inspect their visual selections.

\texttt{get\_\allowbreak{}grasp\_\allowbreak{}contact} reports which fingers contact the environment, their contact impulses in N\,s, and their world-frame poses from forward kinematics. These measurements complement the physical finger gap and RGB observations when the agent assesses contact or object retention. Contact feedback is anonymous: the reported poses locate the finger links rather than exact surface-contact points, and the agent must infer what was touched.
\begin{table}[!htb]
\centering\footnotesize
\setlength{\tabcolsep}{3pt}
\renewcommand{\arraystretch}{1.12}
\caption{\textbf{Robot-state measurements and RGB observations.} State queries and camera tools provide feedback for spatial estimation and action assessment. Contact measurements identify the contacting fingers but not the contacted objects.}
\label{tab:api-observation}
\begin{tabularx}{\linewidth}{@{}>{\raggedright\arraybackslash}p{.29\linewidth}>{\raggedright\arraybackslash}p{.29\linewidth}>{\raggedright\arraybackslash}X@{}}
\toprule
Tool & Agent-supplied inputs & Returned information \\
\midrule
\texttt{get\_\allowbreak{}world\_\allowbreak{}frame} & None & World-axis directions and coordinate conventions. \\
\texttt{get\_\allowbreak{}embodiment} & None & Robot dimensions, gripper geometry, and TCP/camera-mount conventions. \\
\texttt{get\_\allowbreak{}camera\_\allowbreak{}info} & Camera & Camera intrinsics, extrinsics, and image dimensions. \\
\texttt{get\_\allowbreak{}arm\_\allowbreak{}pose} & Arm & Measured end-effector and TCP poses, including orientation axes. \\
\texttt{get\_\allowbreak{}gripper\_\allowbreak{}state} & Arm & Gripper drive readback, physical finger gap, and measurement availability. \\
\texttt{get\_\allowbreak{}robot\_\allowbreak{}state} & Optional arm selection & Combined arm poses, gripper state, contact feedback, and joint state. \\
\texttt{get\_\allowbreak{}grasp\_\allowbreak{}contact} & Arm & Contacting fingers, contact impulses, and finger-link poses. \\
\texttt{capture\_\allowbreak{}head} & None & Workspace RGB overview with observation identifier and calibration. \\
\texttt{capture\_\allowbreak{}wrist} & Active arm and optional views & One or both wrist-camera RGB views with identifiers and calibration. \\
\texttt{capture\_\allowbreak{}evidence\_\allowbreak{}views} & Optional arm and camera views & Selected head/wrist RGB views from the same simulation state. \\
\texttt{draw\_\allowbreak{}marks} & Observation, pixels, and labels & RGB image annotated at the selected pixels. \\
\bottomrule
\end{tabularx}
\end{table}

\begin{figure}[t]
  \centering
  \includegraphics[width=\linewidth]{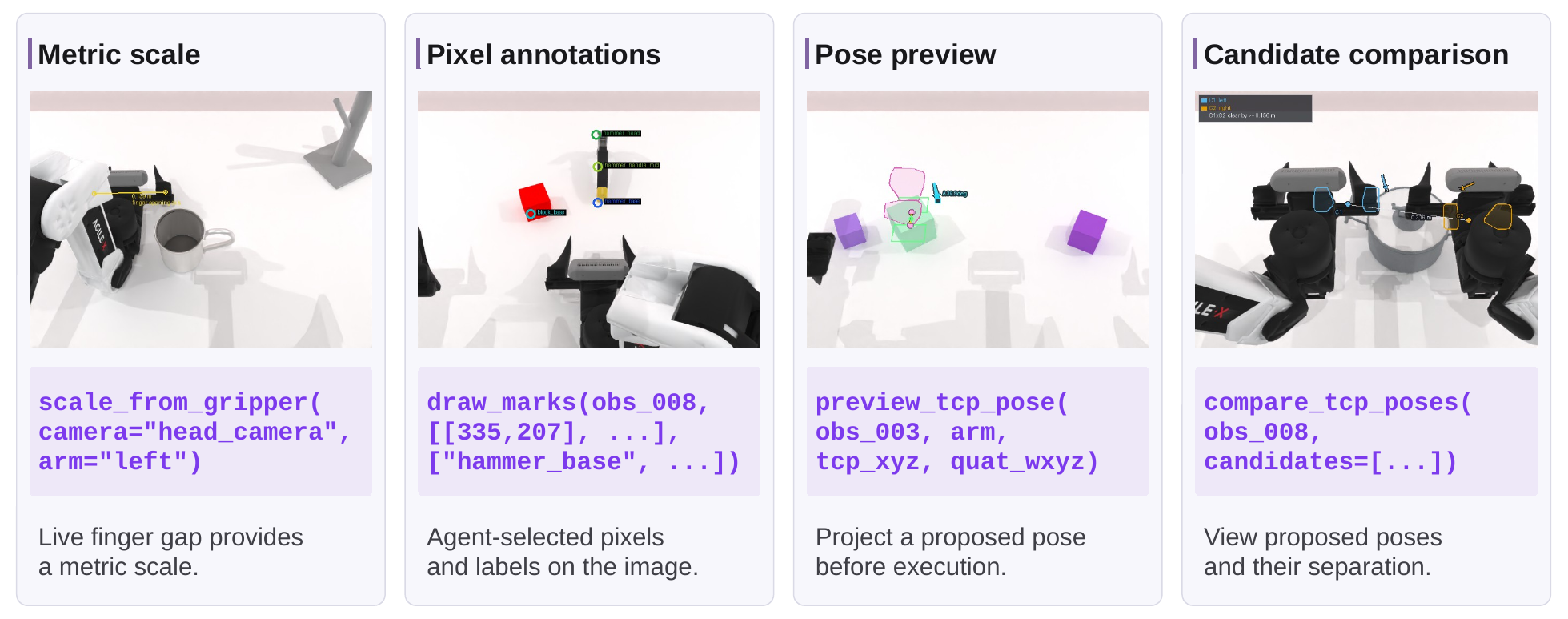}
  \caption{\textbf{Visual feedback for spatial estimation and target construction.} The panels show measured gripper geometry, selected image points, a proposed gripper pose, and a comparison of pose candidates. The agent supplies the image points and proposed poses.}
  \label{fig:visual-workspace}
\end{figure}

\subsection{Spatial calculations and candidate poses}
Spatial estimation combines agent-selected image evidence with calibrated geometry. \texttt{ray} maps a pixel to a world-frame viewing ray, and \texttt{plane\_\allowbreak{}intersect} estimates a 3D point by intersecting that ray with an agent-specified plane. The agent can construct this plane from contact measurements and an assumed surface normal. For triangulation, the agent first selects an arm displacement and invokes \texttt{capture\_\allowbreak{}motion\_\allowbreak{}pair} to acquire wrist images before and after the motion. It then identifies the same scene point in both images and supplies the corresponding pixels to \texttt{triangulate\_\allowbreak{}correspondence}, which estimates the point's 3D position from the measured camera poses. Scale can also be estimated from the projected gripper geometry or an agent-supplied object-size prior. Returned uncertainty reflects the declared pixel, plane, or size uncertainty. \texttt{project} maps a proposed 3D point back into an observation so that the agent can inspect its image alignment.

A grasp target requires a gripper orientation as well as a position. The agent specifies two orthogonal world-frame axes: the approach axis points from the wrist toward the fingertips, and the opening axis follows the direction along which the fingers separate and close. \texttt{grasp\_\allowbreak{}quat\_\allowbreak{}candidates} converts these axes into two quaternion orientations related by a $180^\circ$ wrist rotation. This lets the agent specify the intended grasp geometry without manually deriving quaternions.

Before executing a candidate pose, the agent can inspect its projected gripper geometry with \texttt{preview\_\allowbreak{}tcp\_\allowbreak{}pose} or compare candidate overlays and relative geometry with \texttt{compare\_\allowbreak{}tcp\_\allowbreak{}poses}. \texttt{check\_\allowbreak{}tcp\_\allowbreak{}pose\_\allowbreak{}reachability} then queries cuRobo for a trajectory from the current arm configuration to a proposed TCP pose, returning feasibility and diagnostics without moving the robot. For observation planning, \texttt{camera\_\allowbreak{}aim\_\allowbreak{}pose} computes a TCP pose that directs a wrist camera toward a selected 3D point, accounting for the calibrated camera mount.
\begin{table}[!htb]
\centering\footnotesize
\setlength{\tabcolsep}{3pt}
\renewcommand{\arraystretch}{1.12}
\caption{\textbf{Spatial estimation and target-pose construction.} Tools convert agent-selected pixels, geometric assumptions, and gripper axes into 3D estimates and candidate poses. Projection and reachability checks support inspection before execution.}
\label{tab:api-geometry}
\begin{tabularx}{\linewidth}{@{}>{\raggedright\arraybackslash}p{.29\linewidth}>{\raggedright\arraybackslash}p{.29\linewidth}>{\raggedright\arraybackslash}X@{}}
\toprule
Tool & Agent-supplied inputs & Returned information \\
\midrule
\texttt{ray} & Observation and pixel & World-frame origin and direction of the viewing ray. \\
\texttt{project} & Observation and 3D point & Pixel coordinates of the point in the selected view. \\
\texttt{plane\_\allowbreak{}intersect} & Observation, pixel, plane point/normal, and uncertainty & Estimated 3D intersection and propagated uncertainty. \\
\texttt{capture\_\allowbreak{}motion\_\allowbreak{}pair} & Arm and displacement & Wrist RGB views before/after motion, measured camera displacement, and execution feedback. \\
\texttt{triangulate\_\allowbreak{}correspondence} & Motion pair and corresponding pixels & Estimated 3D point, uncertainty, and triangulation diagnostics. \\
\texttt{scale\_\allowbreak{}from\_\allowbreak{}object\_\allowbreak{}size} & Observation, bounding box, size prior, and extent axis & Coarse depth estimate from the selected image extent and size prior. \\
\texttt{scale\_\allowbreak{}from\_\allowbreak{}gripper} & Camera and arm & RGB image with projected fingertips, their metric separation, and local image scale. \\
\texttt{grasp\_\allowbreak{}quat\_\allowbreak{}candidates} & Gripper approach and opening axes & Two quaternion orientations differing by a $180^\circ$ wrist rotation. \\
\texttt{camera\_\allowbreak{}aim\_\allowbreak{}pose} & Wrist camera, 3D target, and optional pitch/standoff & TCP pose for viewing the target and its predicted image projection. \\
\texttt{preview\_\allowbreak{}tcp\_\allowbreak{}pose} & Observation, arm, TCP pose, and optional finger gap & Gripper geometry overlaid on the image at the proposed pose. \\
\texttt{compare\_\allowbreak{}tcp\_\allowbreak{}poses} & Observation and two to four TCP poses & Candidate overlays and pairwise geometric comparisons. \\
\texttt{check\_\allowbreak{}tcp\_\allowbreak{}pose\_\allowbreak{}reachability} & Arm, target position, and optional orientation & Whether a trajectory was found, with planner diagnostics; no motion is executed. \\
\bottomrule
\end{tabularx}
\end{table}

\noindent\textbf{Observed geometric tool use.} The initial instructions identify camera-motion triangulation, visible gripper geometry, and an agent-supplied size prior as non-contact routes to metric scale, without requiring any one route (Appendix~\ref{app:instructions}). Camera motion can gather a second RGB view without deliberately touching an object; unlike contact probing, this need not disturb the scene during coarse localization. It remains a physical robot motion subject to the same execution safeguards and does not guarantee freedom from unintended contact.
\label{app:policy-estimation}

Astra uses triangulation in all 75 attempts, with 55 successes (73.3\%; Table~\ref{tab:estimation-outcomes}). Opus (Reference) and Opus (Claude Code) invoke triangulation in 10 and 24 attempts, with task success rates of 50.0\% (5/10) and 25.0\% (6/24), respectively. Contact probing is much more prevalent in the two Opus configurations, appearing in 71/75 and 67/75 attempts, compared with 6/75 for Astra.

Geometric computation and contact probing can contribute to the same estimation procedure. A contact measurement, together with an assumed surface normal, can define the reference plane for ray--plane intersection. Across the eight configurations other than Astra, 309/385 attempts using ray--plane intersection also invoke contact probing (80.3\%), compared with 0/54 for Astra. These statistics characterize the use of individual tools within potentially shared estimation procedures.

\begin{table}[!htb]
\centering\footnotesize
\setlength{\tabcolsep}{3pt}
\renewcommand{\arraystretch}{1.12}
\caption{\textbf{Task success associated with geometric and contact tool use.} Entries report task success rates among attempts invoking each tool, with successful/total counts in parentheses. Each configuration has 75 attempts. Usage includes direct calls and calls within programs, counting each attempt once per tool. An attempt may contribute to multiple columns, and contact probing can provide the plane reference used by ray--plane intersection. Triangulation denotes calls to \texttt{triangulate\_\allowbreak{}correspondence}. Bold marks the highest observed rate per column. Task subsets and sample sizes differ across entries.}
\label{tab:estimation-outcomes}
\begin{adjustbox}{max width=\linewidth}
\begin{tabular}{@{}lccccc@{}}
\toprule
& \multicolumn{5}{c}{Task success rate (\%)} \\
\cmidrule(l){2-6}
Configuration & \shortstack[b]{Ray--plane\\intersection} & Triangulation & \shortstack[b]{Gripper\\scale} & \shortstack[b]{Object-size\\prior} & \shortstack[b]{Contact\\probe} \\
\midrule
GPT-6 Astra (Codex CLI) & \textbf{72.2} (39/54) & \textbf{73.3} (55/75) & \textbf{66.7} (4/6) & 66.7 (10/15) & \textbf{83.3} (5/6) \\
Claude Opus 5 (Reference) & 38.1 (8/21) & 50.0 (5/10) & 20.0 (1/5) & 50.0 (1/2) & 50.7 (36/71) \\
Claude Opus 5 (Claude Code) & 29.6 (8/27) & 25.0 (6/24) & 0.0 (0/1) & \textbf{100.0} (2/2) & 46.3 (31/67) \\
Gemini 3.6 Flash (Reference) & 21.2 (14/66) & 0.0 (0/12) & 9.1 (1/11) & 0.0 (0/1) & 22.1 (15/68) \\
GPT-5.6 Sol (Reference) & 13.4 (9/67) & 6.2 (2/32) & 6.2 (1/16) & 8.3 (1/12) & 22.6 (12/53) \\
Qwen 3.8 Max (Reference) & 0.0 (0/35) & 15.0 (3/20) & 0.0 (0/7) & 50.0 (2/4) & 13.5 (10/74) \\
Grok 4.6 (Reference) & 9.2 (6/65) & 4.3 (2/47) & 13.8 (8/58) & 27.8 (5/18) & 20.5 (8/39) \\
Kimi K3 (Reference) & 11.9 (5/42) & 11.8 (2/17) & 7.7 (1/13) & 25.0 (1/4) & 7.4 (4/54) \\
Claude Sonnet 5 (Reference) & 3.2 (2/62) & 0.0 (0/17) & 0.0 (0/15) & 0.0 (0/11) & 2.9 (2/68) \\
\bottomrule
\end{tabular}
\end{adjustbox}
\end{table}

\subsection{Motion and contact feedback}
Motion tools execute agent-specified TCP targets. \texttt{reach\_\allowbreak{}tcp} accepts an absolute position and an optional orientation, while \texttt{move\_\allowbreak{}delta} specifies a displacement from the current TCP position. Their paired variants move both arms in a coordinated call. cuRobo converts these targets into joint trajectories, and the returned robot poses and execution status allow the agent to assess how much of the requested motion occurred, including partial motion before an interruption.

\texttt{probe\_\allowbreak{}contact\_\allowbreak{}along} moves the gripper in bounded increments along an agent-selected direction and checks finger contact after each increment. Starting from free space, it stops when finger contact is detected; more generally, it stops when the contacting fingers change, the travel bound is reached, or execution is interrupted. The measured finger-link poses and known gripper geometry constrain the location of the touched surface. Combined with an assumed surface normal, this provides a reference for defining the plane used in ray--plane intersection to estimate 3D positions.

\texttt{set\_\allowbreak{}gripper} controls the gripper opening with a normalized command from zero (closed) to one (open). The returned physical finger gap and contact feedback help the agent assess the result when an object obstructs closure. Contact conditions can be adjusted through gripper opening and TCP motion; the interface provides position control rather than direct force or impedance control.
\begin{table}[!htb]
\centering\footnotesize
\setlength{\tabcolsep}{3pt}
\renewcommand{\arraystretch}{1.12}
\caption{\textbf{TCP motion, contact probing, and gripper control.} Actions execute agent-specified targets and return measured robot state and execution outcomes, including partial motion before a stop. TCP denotes the tool center point.}
\label{tab:api-motion}
\begin{tabularx}{\linewidth}{@{}>{\raggedright\arraybackslash}p{.29\linewidth}>{\raggedright\arraybackslash}p{.29\linewidth}>{\raggedright\arraybackslash}X@{}}
\toprule
Tool & Agent-supplied inputs & Returned information \\
\midrule
\texttt{reach\_\allowbreak{}tcp} & Arm, target position, and optional orientation & Measured TCP pose and planning/execution outcome. \\
\texttt{reach\_\allowbreak{}both\_\allowbreak{}tcp} & Target positions and optional orientations for both arms & Measured poses of both arms and coordinated execution outcome. \\
\texttt{move\_\allowbreak{}delta} & Arm, world-frame displacement, and optional path mode & Measured TCP pose and execution outcome for the relative motion. \\
\texttt{move\_\allowbreak{}both\_\allowbreak{}delta} & World-frame displacements for both arms & Measured poses of both arms and coordinated execution outcome. \\
\texttt{probe\_\allowbreak{}contact\_\allowbreak{}along} & Arm, direction, travel bound, and step size & Contact change, measured TCP and finger-link poses, and stopping reason. \\
\texttt{set\_\allowbreak{}gripper} & Arm and normalized opening & Drive readback, physical finger gap, contact feedback, and execution outcome. \\
\bottomrule
\end{tabularx}
\end{table}

\subsection{Program composition and termination}
\label{app:api-composition}
\runcode lets the agent combine geometric calculations, observations, and robot actions in a Python program. Conditional branches and loops can use returned measurements to guide subsequent calls, and \texttt{load\_\allowbreak{}image} provides saved RGB images for programmatic analysis. Variables and functions persist across submissions, supporting reuse of estimates and procedures. A timeout or aborted robot action interrupts the program and resets this execution state; the interruption is reported to the agent.

The agent calls \texttt{done} when it judges the task complete or decides it cannot proceed, submitting a report and its Boolean completion claim. The call ends the attempt without revealing the verifier outcome. Recording the agent's judgment separately from the hidden task outcome allows us to evaluate completion assessment, as reported in Appendix~\ref{app:termination}.

\section{Agent Harnesses}
\label{app:harness}
Table~\ref{tab:agent-settings} reports the configurations recorded in the 675 accepted attempts. All use the robot interface in Appendix~\ref{app:api}. The seven reference configurations share the reference harness, using provider-default sampling settings without an explicit temperature override or fixed model seed. Output limits apply per request, with provider-specific accounting of reasoning tokens. Claude Code and Codex CLI manage their own sampling and output limits.

\begin{table}[!htbp]
\centering\footnotesize
\setlength{\tabcolsep}{4pt}
\caption{\textbf{Detailed agent configurations.} Each row contains 75 attempts. Reasoning gives the requested effort; output is the configured per-request token limit. For Claude Code and Codex CLI, the vendor harness manages the output limit. The benchmark sets no additional token cap.}
\label{tab:agent-settings}
\begin{tabularx}{\linewidth}{@{}>{\raggedright\arraybackslash}Xlcr@{}}
\toprule
Configuration & Model identifier & Reasoning & Output tokens \\
\midrule
GPT-6 Astra (Codex CLI) & \texttt{gpt-6-astra} & High & Harness-managed \\
Claude Opus 5 (Reference) & \texttt{claude-opus-5} & High & 128,000 \\
Claude Opus 5 (Claude Code) & \texttt{claude-opus-5} & High & Harness-managed \\
Gemini 3.6 Flash (Reference) & \texttt{gemini-3.6-flash} & High & 65,536 \\
GPT-5.6 Sol (Reference) & \texttt{gpt-5.6} & High & 128,000 \\
Qwen 3.8 Max (Reference) & \texttt{qwen3.8-max} & Medium & 131,072 \\
Grok 4.6 (Reference) & \texttt{grok-4.6} & High & 8,192 \\
Kimi K3 (Reference) & \texttt{kimi-k3} & High & 131,072 \\
Claude Sonnet 5 (Reference) & \texttt{claude-sonnet-5} & High & 128,000 \\
\bottomrule
\end{tabularx}
\end{table}

\subsection{Reference harness}
\label{app:reference-harness}

The reference harness follows a common interaction loop, adapting model requests to each provider's API. Request construction applies the reasoning effort and output limits in Table~\ref{tab:agent-settings}, including adaptive thinking for Opus and Sonnet and thinking preservation for Qwen. Effort levels follow each provider's definitions, so matching labels do not imply equal reasoning-token budgets.

Within each model response, tool calls execute sequentially in the proposed order. Invalid arguments return error feedback, allowing correction in a later turn. A recoverable action interruption cancels the remaining calls from that response, including when the interruption occurs inside a program. Returning the achieved robot state and interruption details lets the agent revise its next action using the changed scene. Recovery decisions therefore remain with the agent.

Context management follows the retention limits in Section~\ref{sec:harness}. Keeping observation identifiers in the text history allows retrieval of earlier images after their removal from the active visual context. Near the context limit, fixed extraction rules reduce older interactions to tool-call records containing selected arguments and returned measurements, including requested targets, achieved poses, contact feedback, and execution outcomes. This reduction removes older assistant prose and program source from the active context without generating a new language-model summary. Further reduction prioritizes the initial instructions and newest interaction, ending the attempt if the request still exceeds the available context.

Transient provider failures trigger bounded retries of the model request before tool execution. Retrying a request leaves the robot state unchanged and does not repeat an executed action.

\subsection{Vendor harnesses}
\label{app:vendor-harness}
Claude Code 2.1.212~\citep{anthropicClaudeCode} and Codex CLI 0.154.0~\citep{openaiCodexCLI} connect to the shared robot API through the Model Context Protocol (MCP)~\citep{modelContextProtocol}. Each harness constructs its own model requests and manages the context, including images. Their strategies for context management therefore differ from the reference harness described above.

Claude Code uses the benchmark tools and \texttt{ToolSearch} for loading tool definitions. The evaluation disables its built-in file, shell, web, and delegation tools. Codex CLI also supports computation and tool-call composition through its built-in execution environment, operating in a read-only sandbox with web search disabled. Neither configuration can access simulator source, privileged object states, or verifier outcomes (Appendix~\ref{app:access}).

Benchmark tool calls follow the shared budget rules in Appendix~\ref{app:budgets}, including calls issued through a vendor harness's computation environment. Loading tool definitions and performing computation alone do not consume this budget. Results characterize each model together with its harness, including differences in context management and available computation tools.

\subsection{Initial prompt}
\label{app:instructions}
Initial instructions describe the robot role, available evidence, program use, budgets, and termination. Task goals and tool schemas provide further details. The excerpts below quote the initial prompt, with punctuation adjusted for readability. Bracketed ellipses mark omissions. FK denotes forward kinematics.

\begin{quote}
You control a dual-arm robot through the provided benchmark tools. [\ldots] Facts: there is no depth sensor and no object ground truth; any metric value you use must come from your own tool evidence. [\ldots] Tool results are typed; \texttt{achieved} can differ from \texttt{commanded} --- trust \texttt{achieved}.

Direct calls and run\_code are equally supported; choose whichever interface fits the step. [\ldots] Three independent non-contact metric-scale sources are available: calibrated camera motion (\texttt{capture\_\allowbreak{}motion\_\allowbreak{}pair} then \texttt{triangulate\_\allowbreak{}correspondence}
), visible FK gripper geometry (\texttt{scale\_\allowbreak{}from\_\allowbreak{}gripper}), and a caller-supplied object-size prior (\texttt{scale\_\allowbreak{}from\_\allowbreak{}object\_\allowbreak{}size}, always coarse). These obtain metric evidence without deliberate scene contact.

\end{quote}

\noindent\textbf{Records for analysis.}
\label{app:recording}
The 675 attempts record model responses, programs, tool results, RGB observations, and execution videos. Actions within programs keep their order and association with the corresponding charged call. Simulator state and verifier outcomes are stored separately for offline analysis.

\subsection{Program use and execution}
\label{app:policy-programs}

Table~\ref{tab:policy-programs} compares program use and execution errors across configurations. Program share measures the fraction of charged tool calls submitted through \runcode. Program errors count submissions returning Python exceptions or sandbox rejections. Appendix~\ref{app:motion-outcomes} reports robot-action interruptions separately.

The four most successful configurations submit 85.6--92.6\% of charged tool calls through \runcode, compared with 15.6--35.2\% for the remaining configurations (Table~\ref{tab:policy-programs}). Astra combines the highest program share with the lowest program error rate (0.46\%). Together, these results suggest that stronger agents better understand the robot tools and compose their operations into effective manipulation policies.

\begin{table}[H]
\centering\small
\caption{\textbf{Program use and execution errors over 75 attempts per configuration.} Program share is the proportion of charged calls submitted through \runcode. Errors include Python exceptions and sandbox rejections; each entry gives the count over returned programs and its rate. Bold marks the highest program share and lowest error rate.}
\label{tab:policy-programs}
\begin{tabularx}{.77\linewidth}{>{\raggedright\arraybackslash}Xrr}
\toprule
Configuration & \shortstack{Program share (\%)} & \shortstack{Program errors count (\%)} \\
\midrule
GPT-6 Astra (Codex CLI) & \textbf{92.6} & 10/2,167 (\textbf{0.46\%}) \\
Claude Opus 5 (Reference) & 85.6 & 65/2,316 (2.81\%) \\
Claude Opus 5 (Claude Code) & 85.7 & 76/2,387 (3.18\%) \\
Gemini 3.6 Flash (Reference) & 87.4 & 159/2,893 (5.50\%) \\
GPT-5.6 Sol (Reference) & 18.9 & 30/784 (3.83\%) \\
Qwen 3.8 Max (Reference) & 35.2 & 91/1,020 (8.92\%) \\
Grok 4.6 (Reference) & 15.6 & 26/706 (3.68\%) \\
Kimi K3 (Reference) & 17.2 & 47/700 (6.71\%) \\
Claude Sonnet 5 (Reference) & 25.3 & 26/1,130 (2.30\%) \\
\bottomrule
\end{tabularx}
\end{table}

\section{Outcomes and Resource Use}
\label{app:process}

\subsection{Task coverage across three attempts}

\label{app:consistency}
\begin{table}[!htb]
\centering
\footnotesize
\setlength{\tabcolsep}{4pt}
\caption{\textbf{Attempt success and task coverage across three attempts.} Each configuration makes three attempts on each of 25 fixed task instances. Success counts successful attempts, and Coverage counts tasks solved at least once. Columns 0/3--3/3 count tasks by their number of successful attempts.}
\label{tab:consistency}
\begin{tabularx}{.8\linewidth}{>{\raggedright\arraybackslash}Xrrrrrr}
\toprule
Configuration & \shortstack{Success (/75)$\downarrow$} & \shortstack{Coverage (/25)} & \shortstack{0/3} & \shortstack{1/3} & \shortstack{2/3} & \shortstack{3/3} \\
\midrule
GPT-6 Astra (Codex CLI) & \textbf{55} & \textbf{22} & 3 & 4 & 3 & \textbf{15} \\
Claude Opus 5 (Reference) & 37 & 19 & 6 & 8 & 4 & 7 \\
Claude Opus 5 (Claude Code) & 34 & 14 & 11 & 3 & 2 & 9 \\
Gemini 3.6 Flash (Reference) & 15 & 8 & 17 & 4 & 1 & 3 \\
GPT-5.6 Sol (Reference) & 12 & 9 & 16 & 7 & 1 & 1 \\
Qwen 3.8 Max (Reference) & 11 & 6 & 19 & 3 & 1 & 2 \\
Grok 4.6 (Reference) & 9 & 6 & 19 & 4 & 1 & 1 \\
Kimi K3 (Reference) & 9 & 5 & 20 & 1 & 4 & 0 \\
Claude Sonnet 5 (Reference) & 2 & 2 & 23 & 2 & 0 & 0 \\
\bottomrule
\end{tabularx}
\end{table}
Table~\ref{tab:consistency} separates tasks solved at least once from those solved in all three attempts on the same fixed scene.

\subsection{Task-level outcomes}
\begin{figure}[!t]
  \centering
  \includegraphics[width=\linewidth,trim=0 30bp 0 0,clip]{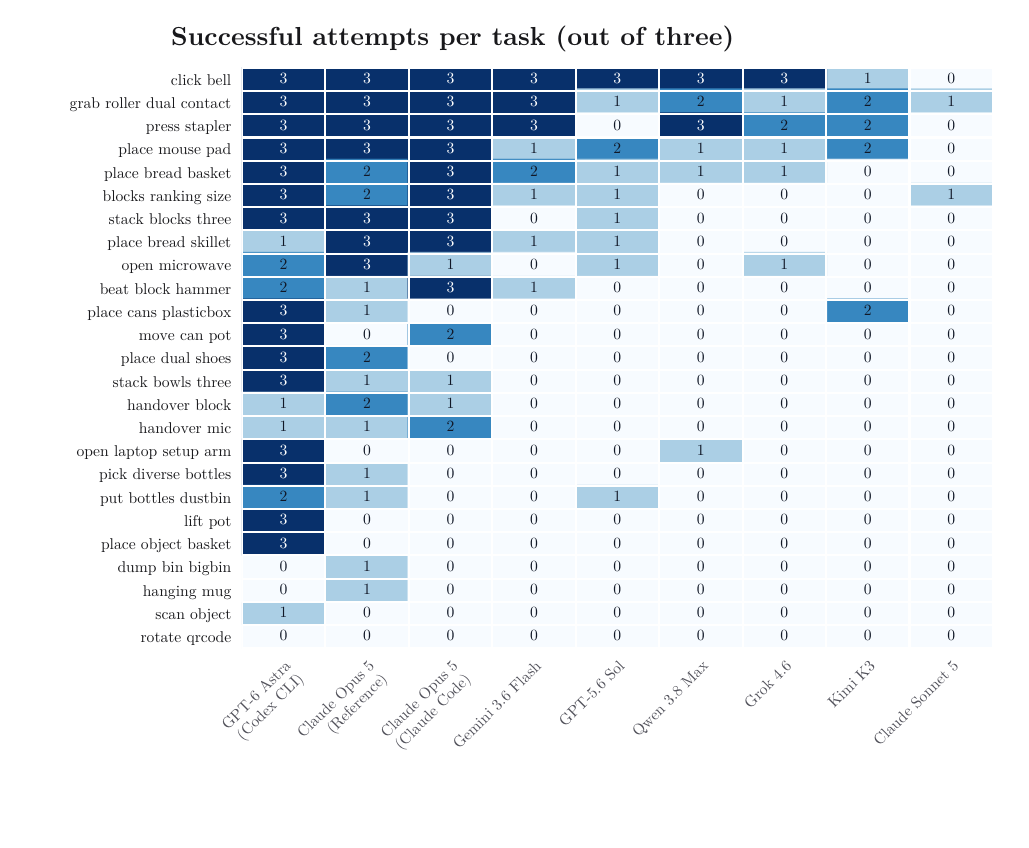}
  \caption{\textbf{Task-level success across nine agent configurations.} Each cell reports successful attempts out of three for one task and configuration. All configurations use the same fixed scene for each of the 25 tasks. Configurations without a harness label use the reference harness.}
  \label{fig:matrix}
\end{figure}

Figure~\ref{fig:matrix} compares success counts over three attempts on each of the 25 tasks. Astra alone succeeds on pot lifting (3/3), whereas Opus (Reference) alone succeeds on bin emptying and mug hanging (1/3 each). These contrasts reveal task-specific strengths beyond the overall ranking.

\subsection{Cost and elapsed time}
\label{app:cost}
Table~\ref{tab:main} reports resource use over all attempts. Simulated time measures physical execution. Wall time also includes inference, provider waiting, and tool processing. Inference cost is expressed in API-equivalent USD.
\begin{table}[!htb]
\centering
\footnotesize
\setlength{\tabcolsep}{4pt}
\caption{\textbf{Resource use over 75 attempts per configuration.} Calls are averaged per attempt; simulated, wall, and provider-wait times are medians. Costs are API-equivalent USD, with median and interquartile range (IQR) computed per attempt. Waiting time is available only for the reference harness. Bold marks the lowest total cost and median wall time.}
\label{tab:main}
\begin{tabularx}{.9\linewidth}{>{\raggedright\arraybackslash}Xrrrrrrr}
\toprule
 & \multicolumn{1}{c}{Calls} & \multicolumn{3}{c}{Median time (s)} & \multicolumn{3}{c}{Inference cost (USD)} \\
\cmidrule(lr){2-2}\cmidrule(lr){3-5}\cmidrule(lr){6-8}
Configuration & Mean & Sim. & Wall & Wait & Total & Median & IQR \\
\midrule
GPT-6 Astra (Codex CLI) & 31.2 & 32.7 & \textbf{477.2} & -- & 252.87 & 3.09 & 2.06--4.35 \\
Claude Opus 5 (Reference) & 36.1 & 45.3 & 1,193.1 & 910 & 297.84 & 3.71 & 2.48--5.29 \\
Claude Opus 5 (Claude Code) & 37.1 & 57.7 & 1,359.4 & -- & 380.09 & 4.88 & 3.34--6.35 \\
Gemini 3.6 Flash (Reference) & 44.1 & 45.0 & 781.6 & 501 & \textbf{52.00} & 0.61 & 0.36--0.88 \\
GPT-5.6 Sol (Reference) & 55.2 & 28.2 & 636.4 & 447 & 259.85 & 3.04 & 2.10--4.54 \\
Qwen 3.8 Max (Reference) & 38.6 & 61.3 & 1,949.4 & 1,620 & 115.27 & 1.38 & 0.95--1.92 \\
Grok 4.6 (Reference) & 60.4 & 17.8 & 952.4 & 818 & 150.73 & 1.73 & 1.31--2.34 \\
Kimi K3 (Reference) & 54.4 & 39.2 & 2,727.6 & 2,372 & 231.65 & 2.81 & 1.64--3.93 \\
Claude Sonnet 5 (Reference) & 59.6 & 29.5 & 1,058.9 & 845 & 143.84 & 1.79 & 1.35--2.38 \\
\bottomrule
\end{tabularx}
\end{table}

API-equivalent costs apply fixed token rates to all configurations, with identical rates for both Opus harnesses. Uncached input/output rates (USD per million tokens) are Astra 10/50, Opus 5/25, Sonnet 2/10, GPT-5.6 Sol 4/20, Gemini 0.75/3.75, Qwen 2/6, Kimi 1.2/4.8, and Grok 2/6.

Calculations account separately for cache reads and writes, apply request-level long-context pricing, and count reasoning tokens once. Claude Code output usage combines the logged thinking-token estimate with a response-length estimate at four UTF-8 bytes per non-thinking token.

\section{Policy Execution Analysis}
\label{app:construction}

\subsection{Checkpoint annotation and replay validation}
\label{app:policy-checkpoints}
\label{app:checkpoint-references}
\label{app:checkpoint-evidence}
Synchronized TCP and object poses from successful expert executions~\citep{chen2025robotwin2} provide the TCP-to-object offsets for spatial annotation (Section~\ref{sec:construction}). Placement offsets use frames before release. Container dimensions and bowl rims determine the region shapes. Calibration uses pooled successful attempts to account for alternative grasp locations and placement choices. Each checkpoint then uses the same criteria, tolerances, and dependencies across all configurations and replays.

\noindent\textbf{Matching and timing.} Task success provides evidence for a subgoal only when the verifier's success conditions imply that subgoal. When a TCP record gives position bounds, a spatial match requires the full bounded region to lie inside the checkpoint region. References that follow moving objects use TCP and object poses from the same time. The match time is the first call with supporting evidence, which can occur after physical attainment. Matches supported only by evidence at termination receive the final call index.

\noindent\textbf{Replay validation.} Replays follow the recorded calls and programs while logging TCP and object states. Validating an entire attempt requires agreement with the original record on tool-call and action order, execution statuses, physical-step counts, and the final verifier outcome. Available TCP endpoints and object positions, including initial positions, are required to agree within 5 mm. The comparison accounts for recorded cancellations and action returns interrupted at termination. Any unexplained missing action invalidates the replay. This 5 mm replay tolerance is separate from the spatial checkpoint tolerances.

A replay with incomplete validation or a different final outcome can still provide an endpoint for an individually verified action. Verification checks execution order, physical-step counts, and available original TCP and object positions under the same criteria. These endpoints can establish spatial matches, but cannot change the original task outcome or establish a final subgoal without evidence from the original attempt. All original positive matches and all 675 attempts remain in the analysis.

\subsection{Attempt-level checkpoint progress}
\label{app:checkpoint-aggregation}
\begin{figure}[!t]
\centering
\includegraphics[width=\linewidth]{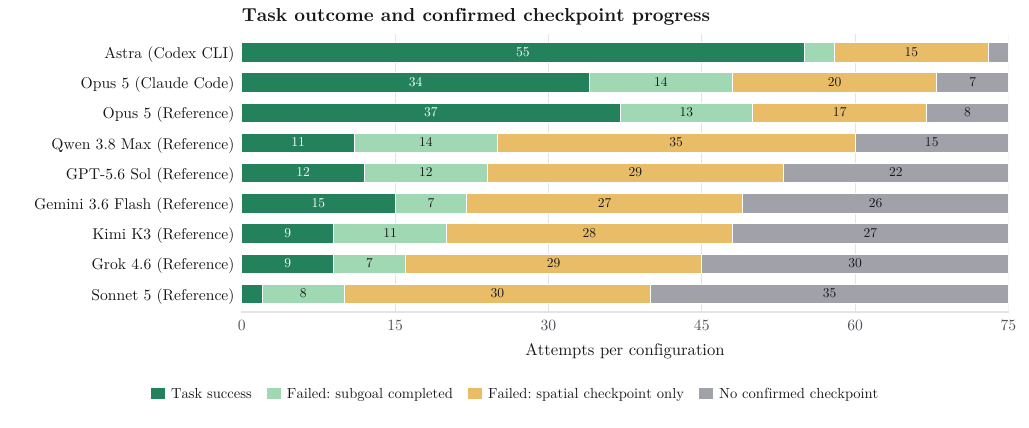}
\caption{\textbf{Task outcomes and confirmed checkpoint progress.} Each configuration has 75 attempts. The bars separate task success from failed attempts with at least one subgoal attained, spatial matches only, or no confirmed checkpoint. Configurations are ordered by the number of attempts with task success or at least one confirmed checkpoint.}
\label{fig:partial-execution}
\end{figure}

Figure~\ref{fig:partial-execution} reports how many attempts achieve task success or at least one checkpoint. Of the 491 failed attempts, 89 satisfy a subgoal during execution, 230 have spatial matches only, and 172 have no confirmed checkpoint. Classification combines the original task outcomes with checkpoint records supplemented by validated replays. Attempts in the last group may still contain motion outside the checkpoint regions.

Task success or checkpoint progress occurs in 73/75 attempts for Astra, 68/75 for Opus (Claude Code), and 67/75 for Opus (Reference). The two Opus configurations thus have similar rates of confirmed progress, although Reference completes more attempts (37 versus 34). Appendix~\ref{app:failure-analysis} examines where execution fails within these attempts.

\subsection{Checkpoint catalogs}
Tables~\ref{tab:subgoal-catalog} and~\ref{tab:checkpoint-catalog} list the subgoal criteria and spatial rules, using the task identifiers in Table~\ref{tab:tasks}. The subgoal catalog lists component conditions before their combinations. A composite checkpoint requires its components to hold simultaneously, so it records a joint outcome rather than a separate action. Independent conditions can occur in any order. The spatial catalog likewise lists approaches before their associated operations, while allowing any order among independent operations. Exact logical expressions and verifier thresholds accompany the analysis data. All distances are in centimetres. Here $\Delta x,\Delta y,\Delta z$ denote differences from a reference TCP, $d_{xy}$ and $d_3$ denote planar and spatial distances, and $\Delta\rho$ denotes radial deviation from a reference ring. These annotations apply only to offline analysis and remain hidden from the agent.

\begingroup\footnotesize
\setlength{\tabcolsep}{3pt}
\renewcommand{\arraystretch}{1.13}
\setlength{\LTcapwidth}{\linewidth}
\begin{longtable}{@{}>{\raggedright\arraybackslash}p{.27\linewidth}cc>{\raggedright\arraybackslash}p{\dimexpr.73\linewidth-4em-6\tabcolsep\relax}@{}}
\caption{\textbf{Subgoal checkpoints and annotation counts for all 25 tasks.} Numbered descriptions list all 55 selected subgoals. The count columns give subgoal (SG) and spatial (SP) checkpoints; the spatial total is 72.}\label{tab:subgoal-catalog}\\
\toprule
Task & SG & SP & Selected subgoals \\
\midrule
\endfirsthead
\multicolumn{4}{l}{\textit{Table~\thetable\ continued.}}\\
\toprule
Task & SG & SP & Selected subgoals \\
\midrule
\endhead
\midrule
\multicolumn{4}{r}{\textit{Continued on the next page.}}\\
\endfoot
\bottomrule
\endlastfoot
\texttt{beat\_\allowbreak{}block\_\allowbreak{}hammer} & 1 & 2 & (1) Hammer head aligned with and contacting the block. \\
\texttt{blocks\_\allowbreak{}ranking\_\allowbreak{}size} & 3 & 5 & (1) Large/middle blocks aligned. (2) Middle/small blocks aligned. (3) Both alignments and size order hold together. \\
\texttt{click\_\allowbreak{}bell} & 1 & 1 & (1) Required bell contact with the selected gripper closed, or its retained success event. \\
\texttt{dump\_\allowbreak{}bin\_\allowbreak{}bigbin} & 3 & 2 & (1) Small bin raised to at least 1.0 m. (2) All five balls in the verifier's height band. (3) The lift and all ball-height conditions hold together. \\
\texttt{grab\_\allowbreak{}roller\_\allowbreak{}dual\_\allowbreak{}contact} & 4 & 4 & (1) Left gripper contacts the roller. (2) Right gripper contacts the roller. (3) Roller above 0.80 m. (4) Both closed grippers contact the raised roller. \\
\texttt{handover\_\allowbreak{}block} & 1 & 3 & (1) Moved block's bottom point aligned with and seated on the support's top point. \\
\texttt{handover\_\allowbreak{}mic} & 2 & 2 & (1) Microphone crosses to the receiver side. (2) Microphone above 0.92 m, across the midline, and in gripper contact. \\
\texttt{hanging\_\allowbreak{}mug} & 1 & 2 & (1) Mug functional point meets rack alignment and height conditions. \\
\texttt{lift\_\allowbreak{}pot} & 4 & 4 & (1) Left TCP within 3 cm of its handle. (2) Right TCP within 3 cm of its handle. (3) Pot above 0.82 m. (4) Both handle conditions and upright raised-pot condition hold together. \\
\texttt{move\_\allowbreak{}can\_\allowbreak{}pot} & 2 & 2 & (1) Can beside the pot on its starting side. (2) Side, position, orientation, and set-down height conditions hold together. \\
\texttt{open\_\allowbreak{}laptop\_\allowbreak{}setup\_\allowbreak{}arm} & 2 & 2 & (1) Lid reaches 40\% of its joint range. (2) Lid angle and either-arm lid-edge condition hold together. \\
\texttt{open\_\allowbreak{}microwave} & 1 & 2 & (1) Door reaches 60\% of the upper joint limit. \\
\texttt{pick\_\allowbreak{}diverse\_\allowbreak{}bottles} & 3 & 4 & (1) First bottle at its XYZ target. (2) Second bottle at its XYZ target. (3) Both bottle targets hold together. \\
\texttt{place\_\allowbreak{}bread\_\allowbreak{}basket} & 2 & 3 & (1) First bread piece in its basket region. (2) Second bread piece in its basket region. \\
\texttt{place\_\allowbreak{}bread\_\allowbreak{}skillet} & 2 & 2 & (1) Bread aligned with the pan point and within 6 cm in height. (2) Bread meets that geometry with neither gripper contacting it. \\
\texttt{place\_\allowbreak{}cans\_\allowbreak{}plasticbox} & 2 & 3 & (1) First can within 4 cm in XY of either box point. (2) Second can within 4 cm in XY of either box point. \\
\texttt{place\_\allowbreak{}dual\_\allowbreak{}shoes} & 2 & 3 & (1) Both shoes satisfy an allowed XY assignment. (2) An allowed assignment, both heights, and both orientations hold together. \\
\texttt{place\_\allowbreak{}mouse\_\allowbreak{}pad} & 2 & 2 & (1) Mouse centered within the XY tolerances. (2) Centering and an allowed orientation hold together. \\
\texttt{place\_\allowbreak{}object\_\allowbreak{}basket} & 3 & 4 & (1) Object near and contacting the basket. (2) Basket raised by more than 2 cm. (3) Raised, level basket carries the raised object off the table with contact retained. \\
\texttt{press\_\allowbreak{}stapler} & 1 & 1 & (1) Required stapler contact, or its retained success event. \\
\texttt{put\_\allowbreak{}bottles\_\allowbreak{}dustbin} & 4 & 4 & (1) First bottle in the bin region. (2) Second bottle in the bin region. (3) Third bottle in the bin region. (4) All three bottle conditions hold together. \\
\texttt{rotate\_\allowbreak{}qrcode} & 2 & 2 & (1) Panel meets target orientation. (2) Target orientation and return height hold together. \\
\texttt{scan\_\allowbreak{}object} & 2 & 3 & (1) Object aligned with the scanner axis. (2) Alignment and positive distance below 7 cm hold together. \\
\texttt{stack\_\allowbreak{}blocks\_\allowbreak{}three} & 3 & 5 & (1) Green block aligned above red. (2) Blue block aligned above green. (3) Both stacking relations hold together. \\
\texttt{stack\_\allowbreak{}bowls\_\allowbreak{}three} & 2 & 5 & (1) At least one bowl pair meets the XY alignment condition. (2) All three bowls meet the height-sorted geometric conditions. \\
\end{longtable}
\endgroup

\noindent\textbf{Moving references and relative motion.} Object-relative checkpoints follow the supporting or destination object, using its current pose and a TCP-to-object offset from expert execution. Bowl approaches use an annulus around the current bowl center. Bowl placements use regions around the supporting bowl's rim, while bin transfer uses the container footprint. In ranking and stacking, a freely chosen first placement provides the reference for later placements without counting as a destination checkpoint. The five relative lift checkpoints measure the same arm's vertical displacement from its matched approach, with no XY constraint or upper height limit. Table~\ref{tab:checkpoint-catalog} gives the task-specific rules.

For bread placement in the skillet, the saved asset data permit four possible reconstructions of the skillet reference point. A geometric match counts only when the condition holds under all four reconstructions. Release and contact checks require separate recorded evidence.

\begingroup\footnotesize
\setlength{\tabcolsep}{3pt}
\renewcommand{\arraystretch}{1.13}
\setlength{\LTcapwidth}{\linewidth}
\begin{longtable}{@{}>{\raggedright\arraybackslash}p{.26\linewidth}>{\centering\arraybackslash}p{.07\linewidth}>{\raggedright\arraybackslash}p{.47\linewidth}>{\raggedright\arraybackslash}p{\dimexpr.20\linewidth-6\tabcolsep\relax}@{}}
\caption{\textbf{Spatial rules for all 72 checkpoints.} Distances are in centimetres. Static regions use expert TCP references; moving-reference rules are explained above. L/R denotes arm selection. Prerequisites refer to earlier actions in the same attempt. For relative lifts, $z_{o,0}$ is initial object height and $z_{\rm approach}$ is the same arm's matched approach height.}\label{tab:checkpoint-catalog}\\
\toprule
Checkpoint & Arm & Spatial condition (cm) & Prerequisite \\
\midrule
\endfirsthead
\multicolumn{4}{l}{\textit{Table~\thetable\ continued.}}\\
\toprule
Checkpoint & Arm & Spatial condition (cm) & Prerequisite \\
\midrule
\endhead
\midrule
\multicolumn{4}{r}{\textit{Continued on the next page.}}\\
\endfoot
\bottomrule
\endlastfoot
\addlinespace[3pt]
\multicolumn{4}{l}{\texttt{beat\_\allowbreak{}block\_\allowbreak{}hammer}} \\*
1. Approach hammer & Either & $|\Delta x|\leq 2$, $|\Delta y|\leq 6$, $\Delta z\in[-3,4]$. & -- \\
2. Strike & Either & $d_3\leq 5$. & 1 \\
\addlinespace[3pt]
\multicolumn{4}{l}{\texttt{blocks\_\allowbreak{}ranking\_\allowbreak{}size}} \\*
1. Approach block 1 & Either & $d_{xy}\leq 2$, $\Delta z\in[-3,4]$. & -- \\
2. Approach block 2 & Either & $d_{xy}\leq 2$, $\Delta z\in[-3,4]$. & -- \\
3. Approach block 3 & Either & $d_{xy}\leq 2$, $\Delta z\in[-3,4]$. & -- \\
4. Large/middle placement & Either & Neighbor-relative: ordered X gap $0$--$13$, $|\Delta y|\leq5.5$, $|\Delta z|\leq4$. & Moved object's approach \\
5. Middle/small placement & Either & Neighbor-relative: ordered X gap $0$--$13$, $|\Delta y|\leq5.5$, $|\Delta z|\leq4$. & Moved object's approach \\
\addlinespace[3pt]
\multicolumn{4}{l}{\texttt{click\_\allowbreak{}bell}} \\*
1. Press & Either & $|\Delta x|\leq 6$, $|\Delta y|\leq 6$, $|\Delta z|\leq 4$. & -- \\
\addlinespace[3pt]
\multicolumn{4}{l}{\texttt{dump\_\allowbreak{}bin\_\allowbreak{}bigbin}} \\*
1. Approach deskbin & Either & $d_{xy}\leq 6$, $\Delta z\in[-5,4]$. & -- \\
2. Pour over bin & Either & Translated expert target: $|\Delta x|\leq22.0$, $|\Delta y|\leq32.4$, $|\Delta z|\leq8$ (XY rounded). & 1 \\
\addlinespace[3pt]
\multicolumn{4}{l}{\texttt{grab\_\allowbreak{}roller\_\allowbreak{}dual\_\allowbreak{}contact}} \\*
1. Approach left end & L & $d_{xy}\leq 4$, $|\Delta z|\leq 3$. & -- \\
2. Approach right end & R & $d_{xy}\leq 5$, $|\Delta z|\leq 3$. & -- \\
3. Lift left & L & Same-arm $z-z_{\rm approach}>\max(0,80-z_{o,0})$; XY and upper Z unrestricted. & 1 \\
4. Lift right & R & Same-arm $z-z_{\rm approach}>\max(0,80-z_{o,0})$; XY and upper Z unrestricted. & 2 \\
\addlinespace[3pt]
\multicolumn{4}{l}{\texttt{handover\_\allowbreak{}block}} \\*
1. Approach box & Either & $d_{xy}\leq 7$, $|\Delta z|\leq 4$. & -- \\
2. Handover & Either & $|\Delta x|\leq 10$, $|\Delta y|\leq 6$, $|\Delta z|\leq 6$. & 1 \\
3. Place at support & Either & $d_{xy}\leq 7.5$, $\Delta z\in[-2,5]$. & 2 \\
\addlinespace[3pt]
\multicolumn{4}{l}{\texttt{handover\_\allowbreak{}mic}} \\*
1. Approach microphone & Either & $d_{xy}\leq 3$, $|\Delta z|\leq 3$. & -- \\
2. Handover & Either & $|\Delta x|\leq 12$, $|\Delta y|\leq 6$, $|\Delta z|\leq 6$. & 1 \\
\addlinespace[3pt]
\multicolumn{4}{l}{\texttt{hanging\_\allowbreak{}mug}} \\*
1. Approach mug & Either & $d_{xy}\leq 5$, $|\Delta z|\leq 4$. & -- \\
2. Final hanging target & Either & Rack point: $d_3\leq 3$. & 1 \\
\addlinespace[3pt]
\multicolumn{4}{l}{\texttt{lift\_\allowbreak{}pot}} \\*
1. Approach left handle & L & $d_{xy}\leq 2$, $|\Delta z|\leq 3$. & -- \\
2. Approach right handle & R & $d_{xy}\leq 2$, $|\Delta z|\leq 3$. & -- \\
3. Lift left & L & Same-arm $z-z_{\rm approach}>\max(0,82-z_{o,0})$; XY and upper Z unrestricted. & 1 \\
4. Lift right & R & Same-arm $z-z_{\rm approach}>\max(0,82-z_{o,0})$; XY and upper Z unrestricted. & 2 \\
\addlinespace[3pt]
\multicolumn{4}{l}{\texttt{move\_\allowbreak{}can\_\allowbreak{}pot}} \\*
1. Approach can & Either & $d_{xy}\leq 3$, $|\Delta z|\leq 3$. & -- \\
2. Beside pot & Either & $d_{xy}\leq 7$, $\Delta z\in[-2,5]$. & 1 \\
\addlinespace[3pt]
\multicolumn{4}{l}{\texttt{open\_\allowbreak{}laptop\_\allowbreak{}setup\_\allowbreak{}arm}} \\*
1. Approach laptop & Either & $d_{xy}\leq 5$, $|\Delta z|\leq 4$. & -- \\
2. Open lid & Either & $d_3\leq 6$. & 1 \\
\addlinespace[3pt]
\multicolumn{4}{l}{\texttt{open\_\allowbreak{}microwave}} \\*
1. Approach microwave & Either & $d_{xy}\leq 2$, $\Delta z\in[-6,2]$. & -- \\
2. Open door & Either & $d_3\leq 5$. & 1 \\
\addlinespace[3pt]
\multicolumn{4}{l}{\texttt{pick\_\allowbreak{}diverse\_\allowbreak{}bottles}} \\*
1. Approach bottle 1 & L & $d_{xy}\leq 3$, $|\Delta z|\leq 3$. & -- \\
2. Approach bottle 2 & R & $d_{xy}\leq 3$, $|\Delta z|\leq 3$. & -- \\
3. Hold left & L & $|\Delta x|\leq 10$, $|\Delta y|\leq 10$, $\Delta z\geq -10$. & 1 \\
4. Hold right & R & $|\Delta x|\leq 10$, $|\Delta y|\leq 10$, $\Delta z\geq -10$. & 2 \\
\addlinespace[3pt]
\multicolumn{4}{l}{\texttt{place\_\allowbreak{}bread\_\allowbreak{}basket}} \\*
1. Approach bread 0 & Either & $d_{xy}\leq 2$, $|\Delta z|\leq 3$. & -- \\
2. Approach bread 1 & Either & $d_{xy}\leq 2$, $|\Delta z|\leq 3$. & -- \\
3. Release over basket & Either & $d_{xy}\leq 7$, $\Delta z\in[-2,3]$. & 1 or 2 \\
\addlinespace[3pt]
\multicolumn{4}{l}{\texttt{place\_\allowbreak{}bread\_\allowbreak{}skillet}} \\*
1. Approach bread & Either & $d_{xy}\leq 5$, $\Delta z\in[-3,4]$. & -- \\
2. Bread in pan & Either & Current pan point: $d_{xy}\leq 10$, $|\Delta z|\leq 6$. & 1 \\
\addlinespace[3pt]
\multicolumn{4}{l}{\texttt{place\_\allowbreak{}cans\_\allowbreak{}plasticbox}} \\*
1. Approach object 1 & Either & $d_3\leq 2$. & -- \\
2. Approach object 2 & Either & $d_3\leq 2$. & -- \\
3. Release in box & Either & $d_{xy}\leq 7.5$, $\Delta z\in[-2,5]$. & 1 or 2 \\
\addlinespace[3pt]
\multicolumn{4}{l}{\texttt{place\_\allowbreak{}dual\_\allowbreak{}shoes}} \\*
1. Approach left shoe & Either & $d_{xy}\leq 5$, $\Delta z\in[-3,6]$. & -- \\
2. Approach right shoe & Either & $d_{xy}\leq 5$, $\Delta z\in[-3,6]$. & -- \\
3. Place at shoe box & Either & $d_{xy}\leq 10$, $\Delta z\in[-2,5]$. & 1 or 2 \\
\addlinespace[3pt]
\multicolumn{4}{l}{\texttt{place\_\allowbreak{}mouse\_\allowbreak{}pad}} \\*
1. Approach mouse & Either & $d_3\leq 2$. & -- \\
2. Place at pad & Either & $d_{xy}\leq 4$, $\Delta z\in[-2,5]$. & 1 \\
\addlinespace[3pt]
\multicolumn{4}{l}{\texttt{place\_\allowbreak{}object\_\allowbreak{}basket}} \\*
1. Approach object & Either & $d_3\leq 2$. & -- \\
2. Place in basket & Either & $d_{xy}\leq 7$, $\Delta z\in[-2,4]$. & 1 \\
3. Approach basket & Either & $|\Delta x|\leq 10$, $|\Delta y|\leq 4$, $\Delta z\in[-3,8]$. & -- \\
4. Basket lift & Either & Same-arm $z-z_{\rm approach}>2$; XY and upper Z unrestricted. & 3 \\
\addlinespace[3pt]
\multicolumn{4}{l}{\texttt{press\_\allowbreak{}stapler}} \\*
1. Press & Either & $|\Delta x|\leq 5$, $\Delta y\in[-6,4]$, $\Delta z\in[-3,4]$. & -- \\
\addlinespace[3pt]
\multicolumn{4}{l}{\texttt{put\_\allowbreak{}bottles\_\allowbreak{}dustbin}} \\*
1. Approach bottle 0 & Either & $d_{xy}\leq 4$, $\Delta z\in[-8,6]$. & -- \\
2. Approach bottle 1 & Either & $d_{xy}\leq 4$, $\Delta z\in[-8,6]$. & -- \\
3. Approach bottle 2 & Either & $d_{xy}\leq 4$, $\Delta z\in[-8,6]$. & -- \\
4. Drop over bin & Either & Translated expert target: $|\Delta x|\leq22.0$, $|\Delta y|\leq32.4$, $|\Delta z|\leq8$ (XY rounded). & 1 or 2 or 3 \\
\addlinespace[3pt]
\multicolumn{4}{l}{\texttt{rotate\_\allowbreak{}qrcode}} \\*
1. Approach qrcode & Either & $d_3\leq 2$. & -- \\
2. Return to table & Either & Table-relative $\Delta z\in[-5,6]$; XY unrestricted. & 1 \\
\addlinespace[3pt]
\multicolumn{4}{l}{\texttt{scan\_\allowbreak{}object}} \\*
1. Approach scanner & Either & $d_3\leq 3$. & -- \\
2. Approach object & Either & $d_3\leq 2$. & -- \\
3. Scan alignment & Either & $|\Delta x|\leq 3$, $|\Delta y|\leq 6$, $|\Delta z|\leq 3$. & 1 and 2 \\
\addlinespace[3pt]
\multicolumn{4}{l}{\texttt{stack\_\allowbreak{}blocks\_\allowbreak{}three}} \\*
1. Approach block 1 & Either & $d_3\leq 2$. & -- \\
2. Approach block 2 & Either & $d_3\leq 2$. & -- \\
3. Approach block 3 & Either & $d_3\leq 2$. & -- \\
4. Place green on red & Either & Current support: $|\Delta x|\leq 2.5$, $|\Delta y|\leq 2.5$, $|\Delta z|\leq 4$. & 2 \\
5. Place blue on green & Either & Current support: $|\Delta x|\leq 2.5$, $|\Delta y|\leq 2.5$, $|\Delta z|\leq 4$. & 3 \\
\addlinespace[3pt]
\multicolumn{4}{l}{\texttt{stack\_\allowbreak{}bowls\_\allowbreak{}three}} \\*
1. Approach bowl 1 & Either & Current bowl: $|\Delta\rho|\leq 2$, $|\Delta z|\leq 5$. & -- \\
2. Approach bowl 2 & Either & Current bowl: $|\Delta\rho|\leq 2$, $|\Delta z|\leq 5$. & -- \\
3. Approach bowl 3 & Either & Current bowl: $|\Delta\rho|\leq 2$, $|\Delta z|\leq 5$. & -- \\
4. Two-bowl placement & Either & Support-relative annulus: $|\Delta\rho|\leq2$, $|\Delta z|\leq5$. & Moved object's approach \\
5. Three-bowl placement & Either & Support-relative annulus: $|\Delta\rho|\leq2$, $|\Delta z|\leq5$. & Moved object's approach \\
\end{longtable}
\endgroup

\section{Failure Analysis}
\label{app:failure-analysis}

\subsection{Observed failure stages}
\label{app:task-progress}
\label{app:failure-categories}
For each failed attempt, the failure analysis starts with unmet subgoals whose prerequisites were attained. These include subgoals that held earlier but no longer hold at termination. A completed subgoal needs no further failure analysis, even without a matching spatial checkpoint, because an alternative route may have achieved it. For each unmet subgoal, the analysis follows its spatial checkpoints and object states in action order, including actions within programs. The earliest observed difficulty determines a single category for the attempt.

A missing compatible target request has no event time, so it cannot override an earlier contact or transport difficulty. Without event timing, a category applies only when all remaining evidence supports that category. Different categories with no established order remain unresolved. Stage labels refer to the unmet subgoal. For example, unconfirmed arrival can concern a grasp location or a later placement destination.

\noindent\textbf{Object control and release.} Offline simulator records identify contacts with the task object. Confirming common motion requires at least one sample with both fingers contacting that object during an action. Over the same action, both the TCP and object move at least 3 cm, while the world-frame object-minus-TCP position vector changes by at most 3 cm. These checks track the same object and arm throughout. Confirming object transport to an operation region also requires a spatial match for that arm.

An open-gripper drive state after an action provides release evidence. Transport loss requires earlier common motion, followed by an open-gripper state, TCP displacement of at least 3 cm, and relative drift greater than 3 cm before arrival at the region. After arrival and release, failure to achieve the required object state indicates a subgoal-outcome failure. State loss requires an observed transition from a satisfied to an unsatisfied subgoal. A required event counts as satisfied after its first recorded occurrence and cannot count as state loss.

\begin{figure}[!t]
\centering
\includegraphics[width=0.9\linewidth]{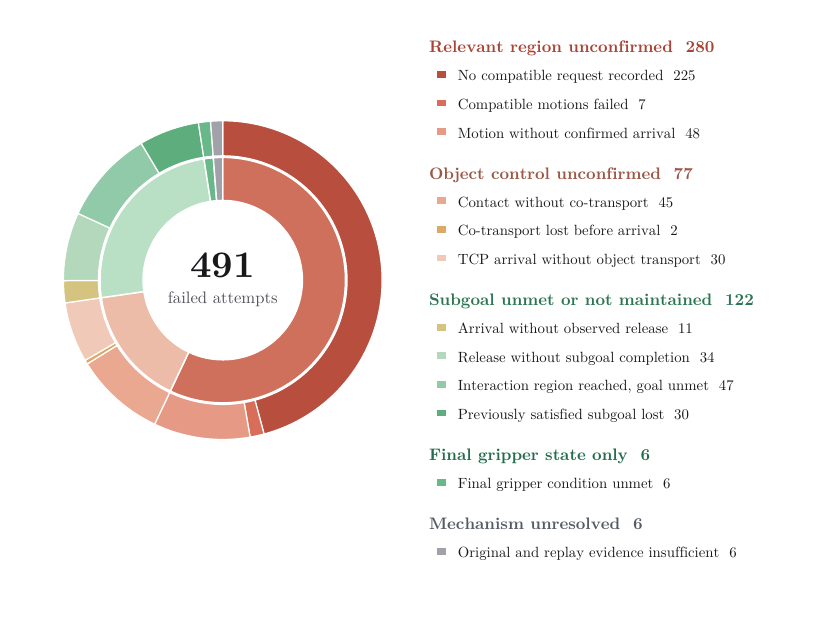}
\caption{\textbf{Observed failure stages and action patterns.} Each of the 491 failed attempts appears once. The inner ring shows the five failure stages. The outer ring divides each stage by the observed action patterns.}
\label{fig:failure-evidence-detail}
\end{figure}

Replays pass full-trajectory validation for 481 of the 491 failed attempts. Original records support a broad failure stage for four more attempts. Six remain unresolved. Figure~\ref{fig:failure-evidence-detail} expands the distribution in Figure~\ref{fig:failure-stages}. Its inner ring shows the same five failure stages, and its outer ring separates the action patterns within each stage.

Unconfirmed arrival covers missing compatible target requests, unsuccessful requests, and motion with no confirmed regional match. Unconfirmed object control covers contact or arrival without confirmed object transport, as well as transport loss before arrival. Subgoal-outcome failures include arrival without the required state and loss of a previously satisfied state. The final-gripper category applies when all selected subgoals hold but the verifier's additional gripper condition remains unmet.

These categories locate difficulties within execution. The underlying checks do not measure grasp force or uniquely identify perception or contact mechanics as the cause. In particular, a missing compatible target request alone does not establish incorrect visual target selection.

\section{Motion Execution and Completion Reports}
The following measures describe robot-action outcomes and the agent's completion judgment. They complement the task-level failure stages above but do not assign additional failure causes.

\subsection{Motion execution outcomes}
\label{app:interface}
\label{app:motion-outcomes}
Table~\ref{tab:motion-failures} counts robot actions that fail to complete, including actions within programs. It reports action execution outcomes separately from task success.

Simulation advances during 2,783 of the 4,045 non-completed actions. A stopped action can therefore change the scene, so subsequent decisions need to account for the achieved state. Planner counts requests that the motion planner rejects. Stall denotes insufficient execution progress, while Deviation denotes excessive departure from the commanded motion. Allowance records stops at limits on motion segments, corrections, or probe travel.
\begin{table}[!htb]
\centering\footnotesize
\setlength{\tabcolsep}{4pt}
\caption{\textbf{Non-completed robot actions.} Each row covers 75 attempts and includes actions within programs. Non-completion reports the number of incomplete actions over all actions, followed by the rate. The remaining columns separate planner refusals, stalls, trajectory deviations, and motion-allowance stops. Bold marks the lowest rate.}
\label{tab:motion-failures}
\begin{tabularx}{\linewidth}{>{\raggedright\arraybackslash}Xrrrrr}
\toprule
Configuration & \shortstack{Non-completion\\count (rate)} & \shortstack{Planner\\count} & \shortstack{Stall\\count} & \shortstack{Deviation\\count} & \shortstack{Allowance\\count} \\
\midrule
GPT-6 Astra (Codex CLI) & 305/2,055 (\textbf{14.8\%}) & 187 & 110 & 3 & 5 \\
Claude Opus 5 (Reference) & 410/2,323 (17.6\%) & 171 & 209 & 6 & 24 \\
Claude Opus 5 (Claude Code) & 444/2,450 (18.1\%) & 191 & 221 & 4 & 28 \\
Gemini 3.6 Flash (Reference) & 665/3,029 (22.0\%) & 400 & 257 & 3 & 5 \\
GPT-5.6 Sol (Reference) & 310/1,611 (19.2\%) & 124 & 166 & 1 & 19 \\
Qwen 3.8 Max (Reference) & 703/2,385 (29.5\%) & 393 & 265 & 3 & 42 \\
Grok 4.6 (Reference) & 207/1,102 (18.8\%) & 68 & 128 & 3 & 8 \\
Kimi K3 (Reference) & 613/1,846 (33.2\%) & 286 & 282 & 7 & 38 \\
Claude Sonnet 5 (Reference) & 388/1,566 (24.8\%) & 202 & 172 & 4 & 10 \\
\bottomrule
\end{tabularx}
\end{table}

\enlargethispage{-8\baselineskip}
\subsection{Completion claims and stopping conditions}
\label{app:termination}
The \texttt{done} report contains the agent's assessment, which is separate from the hidden verifier outcome. Table~\ref{tab:termination} reports stopping conditions and whether an explicit completion claim agrees with that outcome. An absent claim is not a failure judgment.

One of the 322 \texttt{done} reports omits a Boolean claim. Done denotes agent-initiated termination, Tool a charged-call limit, and Phys. a simulated-time limit. The ``Other'' stopping conditions comprise eight wall-time limits and one runtime error.
\begin{table}[!htbp]
\centering
\footnotesize
\setlength{\tabcolsep}{4pt}
\caption{\textbf{Attempt termination and completion claims.} Each configuration has 75 attempts. Correct denotes agreement with the verifier. Over and Under denote incorrect success and failure claims, respectively. Absent indicates no Boolean claim. Bold marks the most frequent stopping condition and completion-claim category within each row.}
\label{tab:termination}
\begin{tabularx}{\linewidth}{>{\raggedright\arraybackslash}Xrrrrrrrr}
\toprule
Configuration & \multicolumn{4}{c}{Stopping condition} & \multicolumn{4}{c}{Completion claim} \\
\cmidrule(lr){2-5}\cmidrule(lr){6-9}
 & Done & Tool & Phys. & Other & Correct & Over & Under & Absent \\
\midrule
GPT-6 Astra (Codex CLI) & \textbf{72} & 0 & 3 & 0 & \textbf{68} & 3 & 1 & 3 \\
Claude Opus 5 (Reference) & \textbf{63} & 0 & 12 & 0 & \textbf{50} & 13 & 0 & 12 \\
Claude Opus 5 (Claude Code) & \textbf{54} & 2 & 19 & 0 & \textbf{42} & 11 & 1 & 21 \\
Gemini 3.6 Flash (Reference) & \textbf{37} & 17 & 21 & 0 & 18 & 19 & 0 & \textbf{38} \\
GPT-5.6 Sol (Reference) & 21 & \textbf{52} & 2 & 0 & 12 & 9 & 0 & \textbf{54} \\
Qwen 3.8 Max (Reference) & 29 & 2 & \textbf{43} & 1 & 23 & 6 & 0 & \textbf{46} \\
Grok 4.6 (Reference) & 20 & \textbf{55} & 0 & 0 & 15 & 5 & 0 & \textbf{55} \\
Kimi K3 (Reference) & 20 & \textbf{36} & 11 & 8 & 13 & 7 & 0 & \textbf{55} \\
Claude Sonnet 5 (Reference) & 6 & \textbf{67} & 2 & 0 & 4 & 1 & 0 & \textbf{70} \\
\bottomrule
\end{tabularx}
\end{table}

Stopping conditions differ with program use (Table~\ref{tab:policy-programs}). The four configurations with the highest program shares reach the tool-call limit in 19/300 attempts (6.3\%), compared with 212/375 (56.5\%) for the other five. Within the latter group, Qwen more often exhausts simulated time (43/75) than tool calls (2/75). Composing several operations in one program saves tool calls, while physical execution still consumes the simulation-time budget.

Astra and both Opus configurations end most attempts with a correct claim. The other six end most attempts without one. Across all nine configurations, 345/491 failed attempts (70.3\%) end without a claim. Gemini also shows that extensive program use does not ensure accurate completion assessment: despite an 87.4\% program share, 19 of its 37 explicit claims incorrectly report success.

\section{Use of Generative AI}
\label{app:ai-use}
We used generative AI tools to aid manuscript drafting and polish writing, search the literature and check citations, edit and debug code, and assist with plotting code and figure and table formatting. We have reviewed all AI-assisted work. We take responsibility for the final content of this work, including text, claims, and artifacts produced with the aid of generative AI.

\end{document}